\documentclass[a4paper,11pt,oneside]{article} 
\usepackage{CS_report} 

\usepackage{twocolceurws}
\usepackage{graphicx}
\usepackage{natbib}
\usepackage{setspace}

\begin{document}

    \twocolumn[{%
      \begin{@twocolumnfalse}

    \captionsetup[figure]{margin=1.5cm,font=small,name={Figure},labelsep=colon}
    \captionsetup[table]{margin=1.5cm,font=small,name={Table},labelsep=colon}

    \begin{center}
        \hrule
        \vskip 8 pt
        \linespread{1.2}\huge {
            Super-Resolution Generative Adversarial Networks based Video Enhancement
        }
        \vskip 8 pt
        \hrule
        \vskip 16 pt
        {\large
        Kağan ÇETİN, Hacer AKÇA\\
        }
        \vskip 4 pt
        {\small Department of Electrical and Electronics Engineering\\ }
        {\small Eskişehir Technical University\\ }
        {\small Eskişehir, Türkiye\\ }
        {\small kagancetin@ogr.eskisehir.edu.tr, hacerakca@ogr.eskisehir.edu.tr\\ }

        \vskip 16 pt
        {\normalsize
            \emph{Special thanks to our supervisor:}\\
        }
        \vskip -12 pt
        {\large
            Prof. Dr. Ömer Nezih GEREK
        }
        
    \end{center}
        
    \vskip 8 pt
    \noindent
\centering
\begin{minipage}{0.9\textwidth}
    \begin{center}
        \vskip 8 pt
        {\large\textbf{Abstract}}
    \end{center}
    \label{ch:abstract}
    \vskip -8 pt
    
    This study introduces an enhanced approach to video super-resolution by extending ordinary Single-Image Super-Resolution (SISR) Super-Resolution Generative Adversarial Network (SRGAN) structure to handle spatio-temporal data. While SRGAN has proven effective for single-image enhancement, its design does not account for the temporal continuity required in video processing. To address this, a modified framework that incorporates 3D Non-Local Blocks is proposed, which is enabling the model to capture relationships across both spatial and temporal dimensions. An experimental training pipeline is developed, based on patch-wise learning and advanced data degradation techniques, to simulate real-world video conditions and learn from both local and global structures and details. This helps the model generalize better and maintain stability across varying video content while maintaining the general structure besides the pixel-wise correctness.
    
    Two model variants—one larger and one more lightweight—are presented to explore the trade-offs between performance and efficiency. The results demonstrate improved temporal coherence, sharper textures, and fewer visual artifacts compared to \textit{traditional} single-image methods. This work contributes to the development of practical, learning-based solutions for video enhancement tasks, with potential applications in streaming, gaming, and digital restoration.
    
    \vskip 8 pt
    \noindent 
    \textbf{Keywords:} Computer vision, Video Super-Resolution, Non-Local Blocks, Artifical Intelligence.
\end{minipage}

    \vskip 26 pt

      \end{@twocolumnfalse}
    }]
    
    
    \section{Introduction}
\label{ch:into} 

\hspace*{6.5mm}The primary objective of this study is to enhance the capabilities of the Super-Resolution Generative Adversarial Network (SRGAN) architectures, mainly designed for Single-Image Super Resolution (SISR) purposes ~\cite{ledigatal}, by extending its adaptability to process three-dimensional (3D) image sequences, thereby leveraging information from consecutive frames to improve resolution—a methodology referred to in the literature as the spatio-temporal approach ~\cite{wang2019edvrvideorestorationenhanced}. Traditionally, SRGAN is designed to operate on individual static images, focusing solely on extracting spatial features to enhance resolution ~\cite{ledigatal}. While effective for single-frame super-resolution, this approach overlooks the temporal relationships inherent in video data, limiting its applicability to dynamic sequences ~\cite{wang2019edvrvideorestorationenhanced} ~\cite{yue2024enhancingspacetimevideosuperresolution}.\newline

This thesis proposes a significant expansion of the SRGAN framework by adapting its input structure from a two-dimensional (2D) ~\cite{dong2015imagesuperresolutionusingdeep}, image-wise configuration to a 3D, frame-sequence-wise format. This modification enables the model to exploit both spatial and temporal information across video frames, offering a more comprehensive solution for video enhancement ~\cite{yue2024enhancingspacetimevideosuperresolution}. By integrating this spatio-temporal perspective, the proposed architecture aims to produce higher-quality outputs with improved consistency and realism compared to conventional SRGAN implementations.\newline

To achieve this, the study investigates the integration of \emph{Non-Local Blocks} ~\cite{wang2018nonlocalneuralnetworks} into a customly designed basic architecture mainly based \emph{Residual Blocks} ~\cite{he2015deepresiduallearningimage}, within a carefully designed \textit{highly} complicated experimental training and evaluation environment. A key innovation explored in this work is the incorporation of spatial feature extraction modules with attention mechanisms, inspired by the seminal paper "Attention Is All You Need" (Vaswani et al., 2017) ~\cite{vaswani2023attentionneed}, which introduced transformative concepts in feature extraction. Specifically, this thesis brings a new approach to the training environment of the model whereas it also adapts non-local blocks—known for their ability to capture long-range dependencies—into the (Single-Image) SRGAN framework to enhance spatio-temporal feature extraction ~\cite{wang2018nonlocalneuralnetworks}. The impact of these modifications is rigorously evaluated, and the results are presented to demonstrate their effectiveness.\newline

Through this research, it is aimed to address the limitations of traditional SRGAN models in video super-resolution tasks and contribute to the growing body of knowledge in spatio-temporal modeling. The evaluation encompasses both quantitative metrics and qualitative insights, shedding light on the potential of the enhanced architecture to advance video enhancement techniques ~\cite{johnson2016perceptuallossesrealtimestyle} ~\cite{abrahamyan2022gradientvariancelossstructureenhanced}. This work not only builds upon existing deep learning methodologies but also paves the way for future explorations in the intersection of attention mechanisms and generative adversarial networks using an innovative and highly experimental approach.

\subsection{Background}
\label{sec:into_back}

\hspace*{6.5mm}At first, architecture of a traditional SRGAN architecture is evaluated based on the basic GAN architecture, with small changes made to the overall approach of the system. First SRGAN explorations are captured in the study “Photo-Realistic Single Image Super-Resolution Using a Generative Adversarial Network” in 2017 ~\cite{ledigatal}, and it built the system that is used nowadays as SRGAN models. Basically, in a super-resolution generative adversarial network, the random noise given as the first module is changed with a low-resolution image that will be compared to the original after the generation of the unknown pixels later by the generator architecture ~\cite{dong2015imagesuperresolutionusingdeep} ~\cite{ledigatal}. The logic of the discriminator model is left the identical, so it can keep examining the results by comparing the generated and the high-resolution ground truth, and then this information can be used to penalize the generator module so it can learn to trick the discriminator by generating more realistic and more undistinguishable images ~\cite{Yang_2019}.\newline

A basic SRGAN model consisted only of a generator and a discriminator where the general concept was the same as the basic GAN idea ~\cite{goodfellow2014generativeadversarialnetworks}; a generator creates a high-resolution images from a given low-resolution input, and a discriminator tries to distinguish if it is fake or real, as it is already said ~\cite{ledigatal}. However, the way chosen to approach to the details is the key to obtain healthy working environments that consist of multiple concurrently connected models as in the SRGAN architecture. After generated images are obtained from the generator, it should be penalized accordingly how much the fake is admissible. This system is explored in the paper “Generative Adversarial Networks” for the first time in 2014 is given in the following Fig.~\eqref{fig:gan_arch} ~\cite{goodfellow2014generativeadversarialnetworks} ~\cite{tian2024generativeadversarialnetworksimage}.

\begin{figure}[!ht]
    \centering
    \includegraphics[scale=0.5]{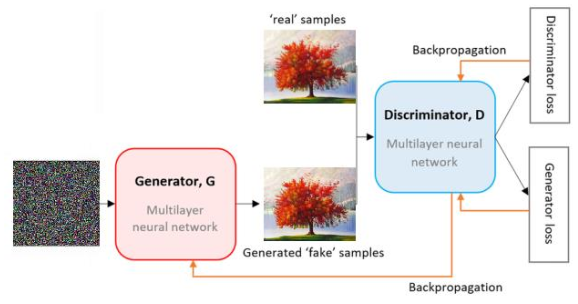}
    \caption{Basic overview of a GAN architecture ~\cite{gan_arch_ref}}
    \label{fig:gan_arch}
\end{figure}

\begin{figure}[!ht]
    \centering
    \includegraphics[scale=0.5]{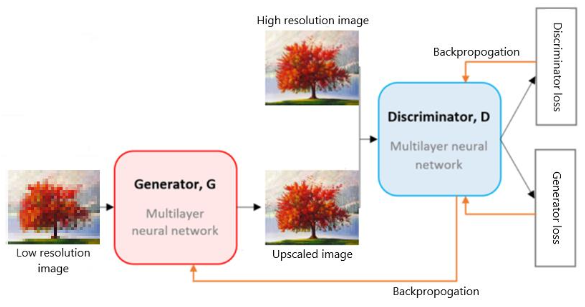}
    \caption{Basic overview of an SRGAN architecture ~\cite{gan_arch_ref}}
    \label{fig:srgan_arch}
\end{figure}

Fig.~\eqref{fig:srgan_arch} shown below also reveals the steps and the approach that should be taken in a healthy SRGAN environment to be able to build a reliable generator with trustworthy results that are desirably similar to the ground truths in the end.\newline

As it is already said, the general architecture of an SRGAN model is being discussed in "Super-Resolution Generative Adversarial Networks (SRGAN)", introduced by Ledig et al. in 2017 ~\cite{ledigatal}, for the first time, and revolutionized image enhancement technologies bu upscaling low-resolution images into perceptually realistic high-resolution outputs. SRGAN’s generator-discriminator architecture, utilizing convolutional layers and residual blocks, prioritizes perceptual quality over pixel-wise accuracy ~\cite{didwania2025laplosslaplacianpyramidbasedmultiscale} ~\cite{johnson2016perceptuallossesrealtimestyle}. However, its original design, optimized for single-image super-resolution, falls short in video enhancement, where temporal dependencies across frames are critical—an area traditional spatial feature extracting modules (i.e. 3D correlation) like \emph{Residual Blocks} do not address ~\cite{yue2024enhancingspacetimevideosuperresolution}.\\

Before expanding SRGAN model's learning capability in spatial domain, it is better take a look at the basic feature-extraction modules that are popular among SRGAN architectures at first.\\

\begin{figure}[!ht]
    \centering
    \includegraphics[scale=0.6]{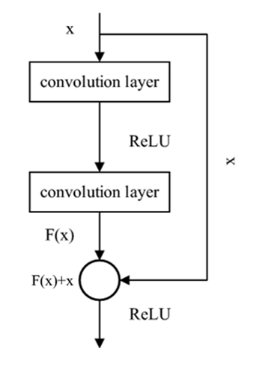}
    \caption{Residual Block by Sum}
    \label{fig:rrdb_bysum}
\end{figure}

Firstly, the residual blocks, split into two different types and it is commonly used in different kinds of image transformation or generation jobs. The first version of residual blocks, which is seen in the \emph{ResNet architecture (He et al., 2015)} ~\cite{he2015deepresiduallearningimage} for the first time, is based on adding the convolved channels into the original image called as residual which is shown below in the Fig.~\eqref{fig:rrdb_bysum}.

\begin{figure}[!ht]
    \centering
    \includegraphics[scale=0.6]{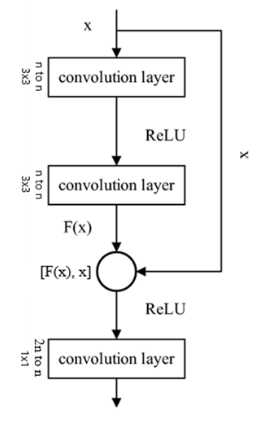}
    \caption{Residual Block by 1x1}
    \label{fig:rrdb_byconv}
\end{figure}

The other residual block approach is done by convolving the output obtained by concatenation of convolved with the original image. In most cases the kernel used to convolve the concatenated is in shape of \(1x1\), which is basically learning its own coefficients to sum up the output obtaiend from the convolution and the original input (i.e. skip connection) instead of adding them directly up. This architecture is seen in the DenseNet architecture for the first time that is shared with public in 2017, and has the following architecture shown in Fig.~\eqref{fig:rrdb_byconv} ~\cite{huang2018denselyconnectedconvolutionalnetworks}.\newline

Also in the following year in 2018, another architecture called as ESRGAN is being publicated and shown the literature that further upgrades can still be made on residual blocks by stacking them consecutively and let them share information among each other. Each output of those blocks are stored and shared directly with the any other residual block waiting on the flow way. The desired architecture is shown in Fig.~\eqref{fig:rrdb_res} and called as \emph{Residual in Residual Dense Block} (RRDB) ~\cite{wang2018esrganenhancedsuperresolutiongenerative}.\newline

Mainly, \emph{Leaky ReLU} is used with a negative slope value of \(0.1\) as activation function which is the function used at the ends of the layers (layers of residual blocks, mainly) to introduce non-linearity to the layers so they can learn complex patterns and correlations.

\begin{equation}
    LReLU(x_i) = max(0.1x_i, x_i)
    \label{eq:lrelu_oneline}
\end{equation}
which is derived from the ordinary ReLU (Rectified Linear Unit)
\begin{equation}
    ReLU(x_i) = max(0, x_i)
    \label{eq:relu_oneline}
\end{equation}

\begin{figure}
    \centering
    \includegraphics[width=0.8\linewidth]{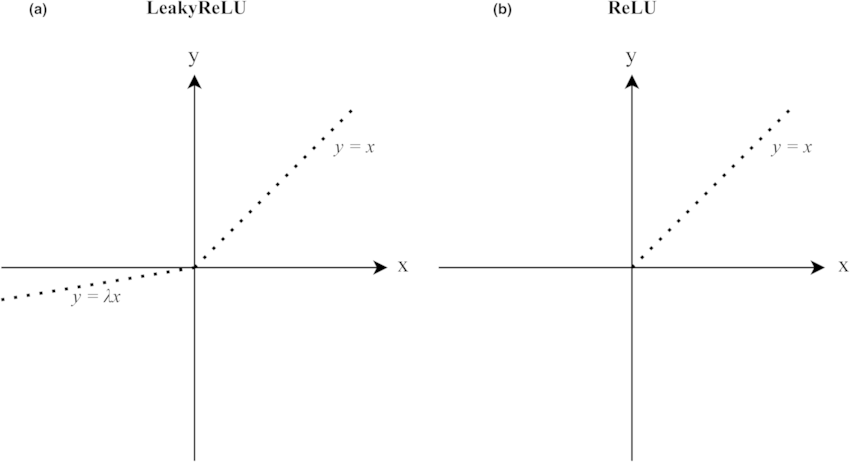}
    \caption{a. Leaky ReLU compared to b. (ordinary) ReLU ~\cite{lrelu_vs_relu_ref}}
    \label{fig:lrelu_vs_relu}
\end{figure}

Their direct illustrations in 2D domain is as follows in Fig.~\eqref{fig:lrelu_vs_relu} ~\cite{xu2015empiricalevaluationrectifiedactivations}\newline

As it is said, this study advances SRGAN into video enhancement by incorporating spatio-temporal feature extraction, culminating in a refined framework. Drawing inspiration from Wang et al. (2018) ~\cite{wang2018nonlocalneuralnetworks}, the final implementation leverages 3D Non-Local Blocks (using dot product as the main operation) to capture inter-frame relationships, moving beyond frame-by-frame processing. Non-local blocks are simply units that computes weighted mean of all pixels in an image batch. Their generic formula is given below in Eq.~\eqref{eq:nlb_generic_form} ~\cite{zhu2021unifyingnonlocalblocksneural}

\begin{equation}
    y_i=1/C(x)\sum\limits_{\forall_j} f(x_i, x_j)g(x_j)
    \label{eq:nlb_generic_form}
\end{equation}

\begin{figure}[!ht]
    \centering
    \includegraphics[scale=0.6]{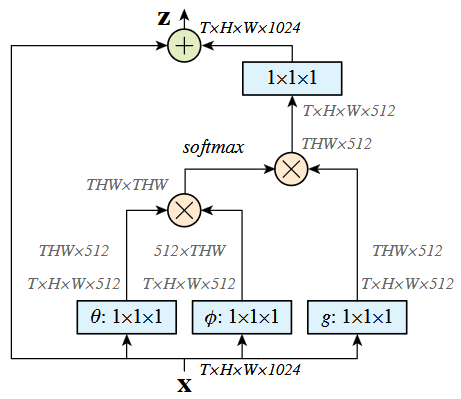}
    \caption{Spacetime View of Non-Local Block ~\cite{wang2018nonlocalneuralnetworks}}
    \label{fig:nlb_diagram}
\end{figure}

The spacetime view (or, diagram) of a non-local block is illustrated below in the Fig.~\eqref{fig:nlb_diagram}.\newline

It is shown that they are more accurate and better in terms of computational efficiency than basic convolutional 3D blocks. Additionally, speaking for \textbf{both} \emph{sequence length} and \emph{(height, width)} pair, formulation of non-local blocks allows usage of \emph{variable} sized inputs, which means usage of \emph{variable} sequence length become possible if the only 3D spatial feature extraction module used is non-local blocks ~\cite{wang2018nonlocalneuralnetworks} ~\cite{zhu2021unifyingnonlocalblocksneural}.\\

Enhanced with effective edge capturing and perceptual loss functions (e.g., \emph{Laplacian} or \emph{Sobel}-based, \emph{Gradient}, \emph{LPIPS}, and \emph{SSIM} losses) ~\cite{abrahamyan2022gradientvariancelossstructureenhanced} \cite{didwania2025laplosslaplacianpyramidbasedmultiscale} ~\cite{johnson2016perceptuallossesrealtimestyle}  ~\cite{zhang2018unreasonableeffectivenessdeepfeatures} ~\cite{nilsson2020understandingssim}, by using a new innovative and experimental \emph{cascading} approach, the model becomes effectively exploiting spatial and temporal contexts concurrently with a desirably good range of evaluation results. Applied to real-world datasets like \textbf{BVI} (specifically, -AOM and -HomTex, thanks to \emph{Dr. Aaron Zhang} from University of Bristol ~\cite{ma2021bvi} ~\cite{nawala2024bvi}), \textbf{Vimeo-90k} and \textbf{REDS} which highly consist of dynamic and realistic scenes, this work evolves SRGAN from its spatial origins into a robust spatio-temporal solution, optimizing video enhancement for practical scenarios.

\subsection{Problem statement}
\label{sec:intro_prob_art}
\hspace*{6.5mm}Video content plays a critical role in numerous domains, including information sharing, entertainment, and scientific analysis in the modern world. However, low-resolution (LR) videos suffer from a loss of visual quality and detail due to limited pixel information, negatively impacting both user experience and the performance of automated analysis systems. Traditional video super-resolution (SR) methods, such as bicubic or bilinear interpolation, often produce blurry, hallucinating, and artificial results, falling short in reconstructing realistic details. The spatio-temporal nature of videos, coupled with the need to maintain consistency and dynamic details in moving scenes, further highlights the limitations of these approaches.  ~\cite{wang2019edvrvideorestorationenhanced}~\cite{yue2024enhancingspacetimevideosuperresolution}\newline

In recent years, deep learning-based methods, particularly Generative Adversarial Networks (GANs), have achieved significant advancements in image super-resolution. For instance, NVIDIA’s Deep Learning Super Sampling (DLSS) technology exemplifies the power of super-resolution in real-world applications. DLSS leverages AI, trained on vast datasets by NVIDIA’s supercomputers, to upscale lower-resolution game frames into high-quality, perceptually realistic outputs, boosting performance while preserving detail. Powered by dedicated Tensor Cores in GeForce RTX GPUs, DLSS has evolved into a cutting-edge solution, with its latest iterations (e.g., DLSS 4, introduced with the RTX 50 Series in 2025) capable of generating up to three additional frames per rendered frame (which is promising to increase FPS (\emph{frames per second}) rate of a video from, say, $\sim$30 FPS to $\sim$120 FPS), enhancing both visual fidelity (usually by 4x scale) and frame rates. This demonstrates super-resolution’s potential as an innovative and scalable technology across dynamic visual media. ~\cite{watson2020deeplearningtechniquessuperresolution}\newline

However, directly applying models like SRGAN to video enhancement remains challenging due to their inability to effectively model spatio-temporal relationships and ensure consistency across consecutive frames. Moreover, existing models are often optimized for static images, neglecting video-specific dynamic features (e.g., motion blur or inter-frame transitions) and real-time processing requirements. This underscores the need for an innovative approach that enhances both visual quality and perceptual realism in video super-resolution. In this thesis, the core problem addressed mainly is "How the super-resolution of video inputs can be improved effectively covering spatio-temporal relationships and preserving inter-frame consistency to achieve perceptually realistic and high-quality results compared to SISR models?". ~\cite{yue2024enhancingspacetimevideosuperresolution}~\cite{dong2015imagesuperresolutionusingdeep} ~\cite{watson2020deeplearningtechniquessuperresolution} \newline

To this end, extending the SRGAN architecture with spatio-temporal feature extraction layers, such as 3D Non-Local Blocks, is proposed as a solution to enhance video enhancement performance ~\cite{wang2018nonlocalneuralnetworks}. This approach could not only advance academic research but also offer practical applications akin to DLSS, such as improving video quality in real-time streaming, gaming, or archival footage restoration ~\cite{watson2020deeplearningtechniquessuperresolution}. However, this extension introduces challenges, including increased model complexity, higher computational costs, and difficulties in generalizing across diverse datasets. This study aims to tackle these challenges firstly by offering a new approach as training environment of the model besides improving the quality of the videos by preserving the spatial coherence besides the temporal consistency, offering both theoretical and practical contributions to the field of video super-resolution, potentially positioning the related approach as a marketable innovation in the growing domain of AI-driven visual enhancement.

\subsection{Aims and objectives}
\label{sec:intro_aims_obj}

\subsubsection{Aims:} 
The primary aim of this thesis is to develop an advanced video super-resolution framework that enhances the quality of low-resolution videos by effectively capturing spatio-temporal relationships and ensuring inter-frame consistency, thereby achieving perceptually realistic and high-quality outputs. Inspired highly by cutting-edge technologies such as NVIDIA’s Deep Learning Super Sampling (DLSS), which leverages super-resolution to transform gaming visuals, this study seeks to extend the capabilities of the Super-Resolution Generative Adversarial Network (SRGAN) architecture for video enhancement. The proposed framework aims to bridge the gap between static image super-resolution and dynamic video processing, offering a solution that is both theoretically robust and practically applicable in domains such as real-time video streaming, gaming, and archival footage restoration. ~\cite{yue2024enhancingspacetimevideosuperresolution} ~\cite{zhu2021unifyingnonlocalblocksneural} ~\cite{watson2020deeplearningtechniquessuperresolution}\\

\subsubsection{Objectives:} To achieve this aim, the following specific objectives have been outlined:

\begin{itemize} 
    \item Enhance the Single-Image SRGAN Architecture by Extending its Dimensions: Extend the existing SRGAN model by integrating spatio-temporal feature extraction layers (e.g., Non-Local Block) to effectively model the temporal dynamics and spatial dependencies inherent in video sequences. ~\cite{wang2018nonlocalneuralnetworks}
    
    \item Improve Perceptual Quality and Consistency: Develop and implement custom loss functions (e.g., Ricker, Gradient, and other perceptual losses) that prioritize perceptual realism and inter-frame coherence, drawing inspiration from DLSS’s ability to produce visually convincing upscaled outputs in real-time gaming environments. ~\cite{johnson2016perceptuallossesrealtimestyle} ~\cite{didwania2025laplosslaplacianpyramidbasedmultiscale}
    
    \item Evaluate Performance Across Datasets: Assess the proposed model’s performance using diverse video datasets (e.g., BVI, Vimeo, REDS) to ensure generalizability and robustness, comparing results against traditional interpolation methods (bicubic, bilinear) and state-of-the-art super-resolution techniques, including DLSS-like benchmarks where applicable.
    
    \item Optimize Computational Efficiency and Develop an Innovative Training Approach: Investigate strategies to balance model complexity and computational cost, enabling the framework to operate efficiently on resource-constrained environments while maintaining high-quality outputs, akin to DLSS’s optimization for real-time performance on RTX GPUs. Check if 16-bit length weights are applicable for the solution, by enabling mixed precision. Additionally, an innovative approach will be developed for training scenario so the model can reach to the desired form faster and learn from different shaped (weight \& height) input scenarios concurrently. ~\cite{watson2020deeplearningtechniquessuperresolution}
    
    \item Demonstrate Practical Applications: Validate the framework’s utility in real-world scenarios, such as enhancing low-quality archival footage or improving video quality in streaming and gaming platforms, positioning it as a potential counterpart to industry innovations like DLSS. Main purpose of this thesis is to adapt the solution into a website designed for this thesis. ~\cite{watson2020deeplearningtechniquessuperresolution}
\end{itemize}
    
By accomplishing these objectives, this research aims to contribute to the advancement of video super-resolution techniques, offering a scalable and innovative solution that aligns with the evolving demands of AI-driven visual enhancement technologies.

\subsection{Solution approach}
\label{sec:intro_sol} 
\textbf{What kind of an experimental approach this thesis recommends?} Following steps were followed in turn at every step:
\begin{itemize} 
    \item Video seqeunces from a desired video dataset (this thesis recommends using one of \emph{Vimeo, REDS, or BVI-AOM}) are split into their frames and loaded partially to the memory iteratively to process batches one-by-one (each batch has a constant length of N frames).
    \item Corresponding image is cropped to a fixed shape, and it is splitted into 16x16 or 32x32 shaped (let's call this \emph{mini-batch split shape}) sub-batches.
    \item Each sub-batch (N many of sequences) is downsampled for a ratio of 2 by using bicubic interpolation method.
    \item Sub-batches that are too dark (detected according to mean of their RGB values) are eliminated since they will affect the model badly in long term training (This can be expanded by checking each side of the images for a fixed length of pixels so it can also be detected if the images have sharp transitions between too dark to normal or vice versa).
    \item Desired gradients are calculated but not back-propogated until whole iteration for each image is done. Each sub-batch can be handled randomly or in turn, also some of them can be eliminated to avoid overfitting, using values >1 for stride.
    \item After the completing the calculation of gradients for each and whole sequence batch, another iteration of the whole process can be repeated by increasing \emph{mini-batch split shape} for x2 or x3.
    \item Same should be applied also for discriminator, keep in mind that both generator and discriminator gradients calculations should be split and not affected by each other.
    \item Using the same batch, an additional iteration can be handled by upscaling the generated batch by a x2 ratio to have a x4 upscaled batch.
\end{itemize}

    \section{Literature Review}
\label{ch:lit_rev} 

\hspace*{6.5mm}Super-resolution (SR) is the process of reconstructing a high-resolution (HR) image or video from a low-resolution (LR) input, aiming to recover details beyond the original data’s limitations. This field has evolved through distinct theoretical stages, each addressing the core challenge of generating realistic, high-quality visuals from sparse information. Early methods, such as the Lanczos filter and Wiener filter, relied on convolution—a mathematical operation that blends local pixel values using a sliding template—to reduce noise and preserve edges during resizing ~\cite{wienerbook} ~\cite{lanczosbook}. However, these techniques were constrained by their fixed rules, struggling to adapt to the complexity of real-world images. Bilinear and bicubic interpolation, also convolution-based, offered simple ways to estimate missing pixels but often produced blurry results, underscoring the need for more adaptable approaches ~\cite{cubicbook} ~\cite{bilinearbicubicdiscussion}.

In the early 2000s, example-based super-resolution emerged as a significant theoretical advance. This method, pioneered by Freeman et al. (2002) ~\cite{firstsr}, used pairs of LR and HR images to learn how to predict fine details from coarse inputs. By employing statistical models like Markov networks, it mapped relationships between LR and HR patches, allowing the system to infer missing information based on learned examples ~\cite{markovmodel}. This marked a shift from rule-based to data-driven approaches, though it required large, diverse datasets to generalize effectively. Around the same time, sparse coding (Yang et al., 2010) introduced a flexible way to represent images using a dictionary of sparse, high-frequency components, enabling more nuanced detail recovery. This method extended convolution’s role by using it to extract and recombine image features dynamically ~\cite{yangmethod}.

Further innovations followed. Neighbor embedding (Chang et al., 2004) adapted manifold learning to super-resolution, projecting LR patches into an HR space while preserving their local geometric relationships ~\cite{changetal}. This approach improved the coherence of reconstructed details but was computationally intensive. Bayesian approaches (Tipping and Bishop, 2003) offered a probabilistic framework, modeling image formation and noise to produce more robust results, especially in cases with uncertain distortions ~\cite{bayessian}. Meanwhile, self-similarity methods (Glasner et al., 2009) exploited repeating patterns within the image itself, eliminating the need for external datasets and leveraging internal structures to enhance resolution ~\cite{glasneretal}. These methods collectively bridged the gap between traditional interpolation and the deep learning era, advancing both theoretical understanding and practical capabilities.

The introduction of deep learning revolutionized super-resolution with the Super-Resolution Convolutional Neural Network (SRCNN) (Dong et al., 2014). SRCNN used convolutional neural networks (CNNs) to learn end-to-end mappings from LR to HR images, dynamically adjusting its filters through training ~\cite{dong2015imagesuperresolutionusingdeep}. This marked a departure from earlier methods, as convolution was now guided by data rather than fixed kernels, significantly improving detail recovery. However, SRCNN’s outputs often lacked natural texture, leading to the development of the Super-Resolution Generative Adversarial Network (SRGAN) (Ledig et al., 2017). SRGAN introduced an adversarial framework, where a generator creates HR images and a discriminator critiques their realism, driving the system toward perceptually convincing results. This approach prioritized visual fidelity over pixel-wise accuracy, aligning enhancement with human perception ~\cite{ledigatal}.

Applying these advancements to video super-resolution presents unique challenges. Videos require not only spatial detail but also temporal consistency across frames. Methods designed for static images often fail to maintain this coherence, resulting in flickering or disjointed motion. In this thesis, all the abilities of the Wiener filter, SRCNN, and 3D Non-Local Blocks will be observable in a singular SRGAN model that is constructed at the end. In this architecture, the Wiener filter stabilizes frames by reducing noise through convolution-based smoothing, while SRCNN refines spatial details using learned convolutional patterns. SRGAN enhances perceptual quality, and 3D Non-Local Blocks capture long-range dependencies across both space and time, ensuring that details remain cohesive throughout the sequence. This combination extends super-resolution from static to dynamic contexts, addressing both spatial richness and temporal flow. ~\cite{dong2015imagesuperresolutionusingdeep} ~\cite{wang2019edvrvideorestorationenhanced} ~\cite{wang2018nonlocalneuralnetworks} ~\cite{wienerbook} 

Convolution remains central across these methods, evolving from fixed templates in traditional approaches to learned, adaptive filters in deep learning. In the constructed model, convolution enables each component—whether stabilizing frames, refining details, or unifying the sequence—to process and blend information effectively. By uniting these approaches, this project overcomes the limitations of earlier methods, such as softened details or broken motion, offering a theoretically grounded solution that treats video enhancement as a unified spatial-temporal process. This contribution has the potential to improve applications like restoring archival footage or enhancing live streams, advancing both the theoretical and practical frontiers of super-resolution. ~\cite{wang2019edvrvideorestorationenhanced} ~\cite{parketal}

\subsection{Spatio-Temporal Learning Architecture built for Video Super-Resolution} 

\begin{figure*}[!ht]
    \centering
    \includegraphics[scale=0.4]{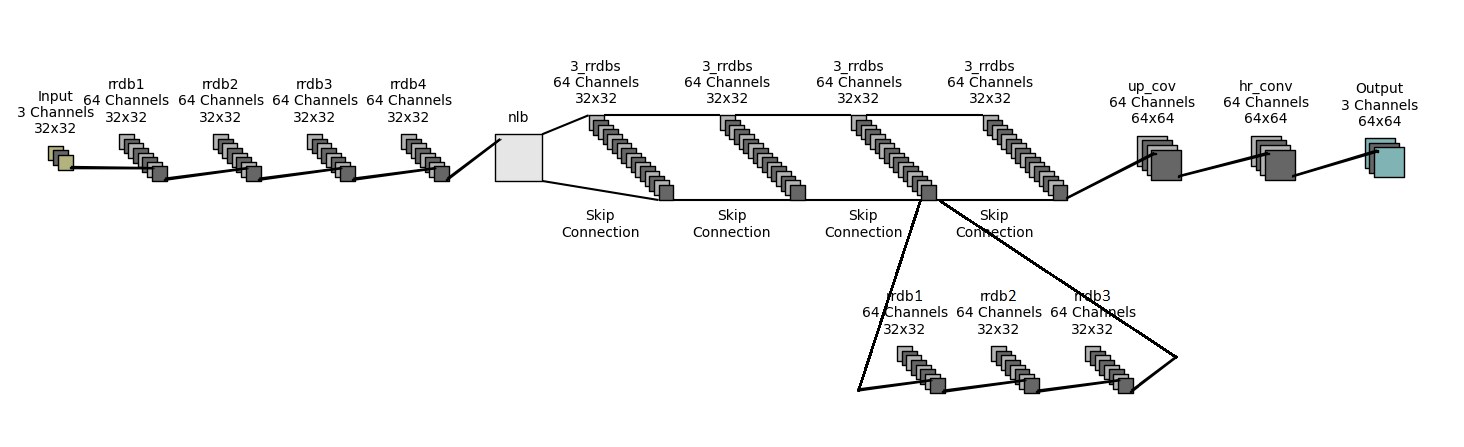}
    \caption{Architecture of \emph{RRDB} based model (has ~27M many of parameters).}
    \label{fig:rrdb_net}
    
    \includegraphics[scale=0.4]{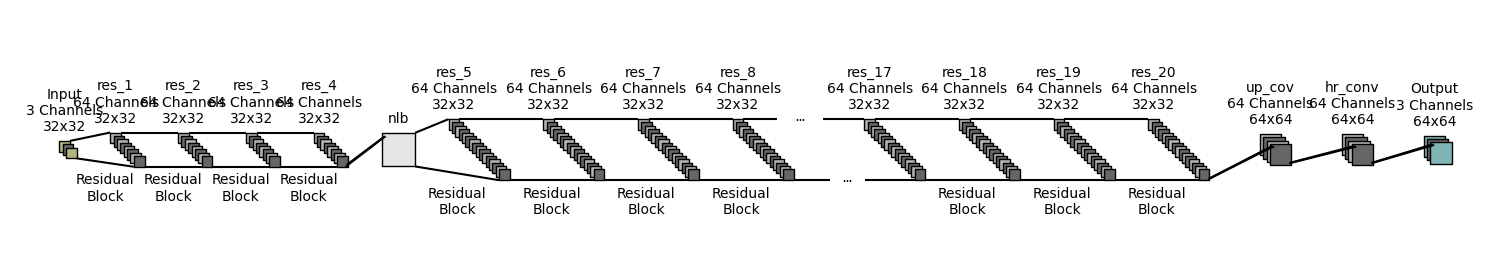}
    \caption{Architecture of \emph{Residual Block} based model (has ~5M many of parameters).}
    \label{fig:res_net}
\end{figure*}

\begin{figure*}[!ht]
    \centering
    \includegraphics[scale=0.6]{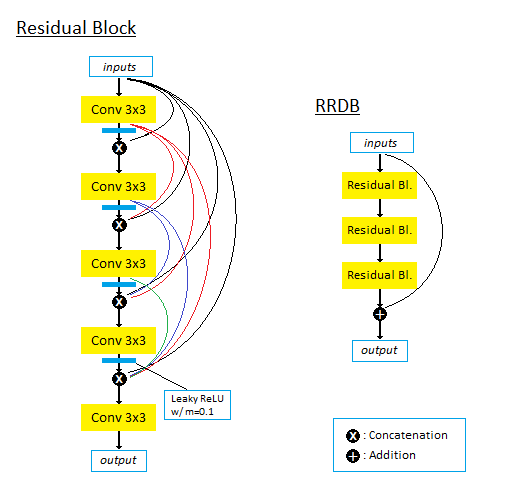}
    \caption{a. \emph{Residual Block} and b. \emph{Residual in Residual Dense Block} modules designed for this project.}
    \label{fig:rrdb_res}
\end{figure*}

\hspace*{6.5mm}At first, it should be noted that two different experimental and basic models for different complexities of jobs are designed. The first shown in Fig.~\eqref{fig:rrdb_net} is built using consecutive RRDBs mainly, whereas the other model which is shown in Fig.~\eqref{fig:res_net} is built based on successive residual blocks. Both of the models can be split into 4 different parts, that they can be called \emph{channel expanding and temporal feature extraction layer, spatial feature extraction layer, reconstruction layer} and \emph{upsampling and channel compression layer}. The \emph{RRDB}s are based on \emph{Residual Blocks}, and each residual block is simply a sequentially connected series of convolutions with skip connections, where flow diagram of those blocks are shown in Fig.~\eqref{fig:rrdb_res}.a and shown in Fig.~\eqref{fig:rrdb_res}.b. In simple terms, each RRDB is also a series (3) of residual blocks with another skip connection at the end.

Classically, residual blocks were implemented based on addition of residuals instead of the method used in this thesis, where the addition process is replaced by concatenation process. In the paper “ESRGAN: Enhanced Super-Resolution Generative Adversarial Networks” (2018) ~\cite{wang2018esrganenhancedsuperresolutiongenerative}, this different and a bit more costly alternative of residual block by sum method has been evaluated and shown to perform even better than the addition alternative of it, which was based on stacking the identity with the processed (convolved) input.

Mathematically, each convolution term is represented as \(y=W*x+b\) where \(W\) is \emph{weight matrix (kernel) of the convolution}, \(*\) denotes \emph{convolution operation}, and \(b\) is the \emph{bias term} ~\cite{wang2018nonlocalneuralnetworks}.

As it is said before, \(LeakyReLU\) is mathematical representation is given in Eq.~\eqref{eq:lrelu}, where \(m\) is specifically set as \(0.1\) for this project:
\begin{equation}
    LeakyReLU(y) = 
    \begin{cases}
      y, & n \geq 0 \\
      -my, &  n < 0
    \end{cases}
    \label{eq:lrelu}
\end{equation}

Assuming \(\sigma (y) = LeakyReLU(y)\), a convolutional operation followed by a leaky relu function in the flow diagram mathematically represents the following Eq.~\eqref{eq:lrelu_and_conv}.

\begin{equation}
    z_i = \sigma (W\ast z_{i-1} + b)
    \label{eq:lrelu_and_conv}
\end{equation}

Assume that the last output we get at the end of a \emph{residual block} is \(Z_{i+5}\), then going from very end to the beginning step by step:
\begin{align}
    Z_i = \sigma (W_i\ast X_i + b_i) \nonumber \\
    = \sigma (Conv_i(X_i)) \nonumber
\end{align}
\begin{align}
    Z_{i+1} = \sigma (W_i\ast (Z_i\mathbin\Vert X_i) + b_{i+1}) \nonumber \\
    = \sigma (Conv_{i+1}(Z_i\mathbin\Vert X_i)) \nonumber
\end{align}
\begin{align}
    Z_{i+2} = \sigma (W_{i+1}\ast (Z_{i+1}\mathbin\Vert Z_i\mathbin\Vert X_i) + b_{i+1}) \nonumber \\
    = \sigma (Conv_{i+2}(Z_{i+1}\mathbin\Vert Z_i\mathbin\Vert X_i)) \nonumber
\end{align}
\begin{align}
    Z_{i+3} = \sigma (W_{i+2}\ast (Z_{i+2}\mathbin\Vert ... \mathbin\Vert Z_i\mathbin\Vert X_i) + b_{i+2}) \nonumber \\
    = \sigma (Conv_{i+3}(Z_{i+2}\mathbin\Vert ... \mathbin\Vert Z_i\mathbin\Vert X_i)) \nonumber
\end{align}
\begin{align}
    Z_{i+4} = \sigma (W_{i+3}\ast (Z_{i+3}\mathbin\Vert ...\mathbin\Vert Z_i\mathbin\Vert X_i) + b_{i+3}) \nonumber \\
    = \sigma (Conv_{i+4}(Z_{i+3}\mathbin\Vert ... \mathbin\Vert Z_i\mathbin\Vert X_i))  \nonumber
\end{align}

In the end, output of the residual block used in this project is as in given in Eq.~\eqref{eq:res_this_project} below.
\begin{equation}
    out_{residual} = \frac{1}{3} Z_{i+5} + \frac{2}{3} X_i
    \label{eq:res_this_project}
\end{equation}

Note the following notations represent the same thing, given in Eq.~\eqref{eq:conv_notations} below.
\begin{equation}
    Conv_{i}(X_i) = W_i\ast X_i + b_{i+1}    
    \label{eq:conv_notations}
\end{equation}

If we go a little further, within the \emph{RRDB} blocks, following mathematical operations are handled:
\[Z_i = Res_i(X_i)\]
\[Z_{i+1} = Res_{i+1}(X_{i+1})\]
\[Z_{i+2} = Res_{i+2}(X_{i+2})\]
\[out_{RRDB} = Z_{i+2} + X_i\]

Correspondingly, RRDB class designed for this project is as follows given in Eq.~\eqref{eq:rrdb_this_project} below.
\begin{equation}
    out_{RRDB} = Res_{i+2}(Res_{i+1}(Res_i(X_i))) + X_i\
    \label{eq:rrdb_this_project}
\end{equation}

On the other hand, \emph{Non-Local Blocks} has a different calculation logic from which its spatial feature extraction ability comes from. Normally, ordinary 2D convolution blocks and its derivations are based on extracting the temporal relationship, which limits the model's learning ability to 2D dataset only. However, with non-local blocks, even in its simplest form, which is called \emph{dot product based non-local blocks}, we get the ability to learn the 3D relationships by calculating the weighted sum of the 3D input sets. Mathematically, it is represented in a very generic form given in Eq.~\eqref{eq:general_nlb_formula} ~\cite{wang2018nonlocalneuralnetworks}.

\begin{equation}
    z_i = W_z\ast y_i+x_i
    \label{eq:general_nlb_formula}
\end{equation}
where the output of a non-local block indeed is as follows in Eq.~\eqref{eq:nlb_formula}
\begin{equation}
    y_i = \frac{1}{C(x)}\sum_{\forall j}f(x_i, x_j)g(x_j)
    \label{eq:nlb_formula}
\end{equation}
\begin{itemize}
    \item \(i\): Index of sequence (in time domain).
    \item \(j\): Index that enumerates all possible positions.
    \item \(f(\cdot)\): A pairwise function f computes a scalar (representing relationship such as affinity) between \(i\) and all \(j\).
    \item \(g(\cdot)\): It is the unary function g (given in Eq.~\eqref{eq:unary_function}) computes a representation of the input signal at the position \(j\).
    \item \(C(\cdot)\): The response is normalized by a factor C(x).
\end{itemize}

The normalization factor \(C(\cdot)\) is given below in Eq.~\eqref{eq:general_norm_c_formula}.
\begin{equation}
    C(x) = \sum_{\forall j} f(x_i, x_j)
    \label{eq:general_norm_c_formula}
\end{equation}

The unary function \(g(\cdot)\) is given below in Eq.~\eqref{eq:general_nlb_formula}.
\begin{equation}
    g(x_j) = W_g\ast x_j
    \label{eq:unary_function}
\end{equation}

Pairwise function \(f(\cdot)\) used in this function (\emph{dot product}) is given below in Eq.~\eqref{eq:dot_product_nlb}.
\begin{equation}
    f(x_i, x_j) = \theta(x_i)^T \phi(x_j)
    \label{eq:dot_product_nlb}
\end{equation}
\begin{itemize}
    \item \(\theta(\cdot)\) and \(\phi(\cdot)\): Two different \emph{1x1} 2D convolution blocks whose calculations result in
    \begin{equation}
        \theta(y_i) = W_{\theta}\ast y_i + b_{\theta}
    \end{equation}
    and
    \begin{equation}
        \phi(y_i) = W_{\phi}\ast y_i + b_{\phi}
    \end{equation}
    \item \(x_i\): The skip connection, thus the residual.
\end{itemize}

Alternatively, additional pairwise functions for non-local blocks exist that can be evaluated later ~\cite{wang2018nonlocalneuralnetworks}, which are:

\begin{itemize}
    \item \textbf{Gaussian:}
    \begin{equation}
        f(x_i, x_j) = e^{x_i^T x_j}
    \end{equation}
    \item \textbf{Embedded Gaussian:}
    \begin{equation}
        f(x_i, x_j) = e^{\theta(x_i)^T \phi(x_j)}
    \end{equation}
    \item \textbf{Dot Product:}
    \begin{equation}
        f(x_i, x_j) = \theta(x_i)^T \phi(x_j)
    \end{equation}
    \item \textbf{Concatenation:}
    \begin{align}
        f(x_i, x_j) = \sigma (w_f^T \cdot \theta(x_i)^T \mathbin \Vert \phi(x_j))
    \end{align}
    \emph{Please note that \(\sigma(\cdot)=ReLU(\cdot)\).}
\end{itemize}

\begin{figure*}[!ht]
    \centering
    \includegraphics[width=0.7\linewidth]{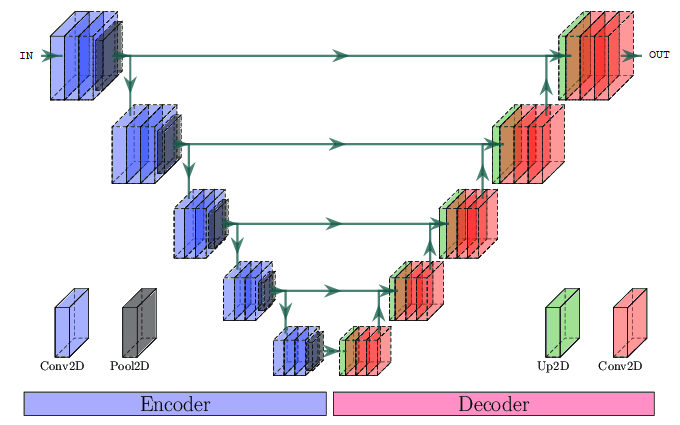}
    \caption{U-Net architecture used as discriminator ~\cite{unet_arch_ref}}
    \label{fig:unet_disc_arch}
\end{figure*}

In this adaptation process, the purpose of the discriminator shall not be changed deeply. Since their performance on image segmentation is proven and it is shown that their encoder \& decoder based architecture is reliably well-performing along large scales of projects ~\cite{ronneberger2015unetconvolutionalnetworksbiomedical}, a U-Net architecture (see Fig.~\eqref{fig:unet_disc_arch}) is built as the discriminator and it can be kept the same as long as it does well in distinguishing artificially made images from real images at first. After a while, it will converge to a non-zero fixed amount of error. Attempting to minimize this error excessively, bring it very close to zero, and significantly increase the complexity of the discriminator block in the process will negatively affect the overall procedure. The backpropagation of a non-zero loss from the discriminator block to the generator is necessary for a healthy process.

Technically, the reason why a U-Net architecture is used as discriminator comes from U-Net architecture's ability to extract a large bandwidth of frequency; thus, high fine details can be extracted, whereas the model also learns to check the structural similarity. Progressive downsampling helps discriminator to capture both local details and global context across multiple scales. This is valuable for distinguishing subtle differences between real and generated HR images, which is a key requirement in SRGANs. Also note that U-Net variants have been used for image-to-image translation showing that they can learn the transitions between two different image types, thus it can also learn the variations between real and fake data. ~\cite{unetproof}\newline

\section{Methodology}
\label{ch:method} 

\subsection{Details of the Patch-Based Training Process}
\label{sect:patch_based_training_proc}

\subsubsection{Data Augmentation}
In the data load part, reliable data augmentation methods have been adapted to the system, which are given below in detail.
\textbf{The first thing}, also the most common, is \emph{random rotation}. A rotation is applied to the batch by a randomly selected amount of degrees from multiples of 90 degrees, which are 0, 90, 180 or 270. \textbf{Secondly}, according to the chance of 50\%, \emph{a flip on the vertical axis} is applied to the batch. \textbf{Finally}, again, according to the chance of 50\%, \emph{a flip on the horizontal axis} is applied to the batch.

\subsubsection{Data Degradation and Data Loading}
After data augmentation, some types of data degradation methods have been applied to the input batch. Firstly, the reason for the application of image degradation techniques is simulating the real-world imperfections and enhancing the model robustness since the inputs are not going to be the exact same ones as those used in the training phase. These techniques illustrate controlled distortions to images, usually related with randomized parameters, enabling models to generalize better to noisy, low-quality, or diverse input data. This subsubsection outlines the set of degradation methods used within the scope of this project.\newline

\noindent
\textbf{Gaussian Blur}
\begin{itemize}
    \item \textbf{What is it?} Gaussian blue applies a gaussian filter to smooth images, reducing high-frequency details. This simulates out-of-focus or low-resolution imaging conditions.
    \item \textbf{How is it implemented?} The input batch is convolved with a Gaussian kernel.
\end{itemize}

\noindent \textbf{Gaussian Noise}
\begin{itemize}
    \item \textbf{What is it?} Gaussian noise introduces random perturbations to pixel values, mimicking sensor noise or graininess in low-light conditions.
    \item \textbf{How is it implemented?} A noise drawn from a Gaussian distribution is added directly to the batch.
\end{itemize}

\noindent \textbf{Contrast and Brightness Adjustment}
\begin{itemize}
    \item \textbf{What is it?} This technique modifies the brightness and contrast of the image to simulate variations in lighting or display conditions.
    \item \textbf{How is it implemented?} Brightness and contrast are adjusted sequentially, where the amplitudes are defined by random in a given range.
\end{itemize}

\noindent \textbf{Frequency-Guided Augmentation}
\begin{itemize}
    \item \textbf{What is it?} Frequency-guided augmentation technique manipulates the batch in the frequency domain using wavelet transforms, altering high-frequency details.
    \item \textbf{How is it implemented?} The batch undergoes wavelet decomposition, followed by random scaling or zeroing of detail coefficients. The modified batch can be reconstructed using inverse wavelet transform, and this is learnable by the model also. ~\cite{freqdomaindataaugmentation} ~\cite{freqdomaindataaugmentation2}
\end{itemize}

\noindent \textbf{CutBlur Mask}
\begin{itemize}
    \item \textbf{What is it?} \emph{CutBlur} applies localized blurring to specific image regions, creating a hybrid of sharp and blurred areas.
    \item \textbf{How is it implemented?} A random mask is created, and the batch is blurred using \emph{average pooling/bilinear downsampling}. ~\cite{yoo2020rethinkingdataaugmentationimage}
\end{itemize}

\noindent \textbf{Diffusion-Based Degradation}
\begin{itemize}
    \item \textbf{What is it?} Diffusion-based degradation applies iterative smoothing to simulate natural degradation processes/\emph{so the Gaussian blur}.
    \item \textbf{How is it implemented?} The batch undergoes iterative Gaussian blurring, with each step applying a convolution operation to diffuse pixel intensities. ~\cite{adaptivedegradation}
\end{itemize}

\noindent \textbf{Content-Aware Degradation}
\begin{itemize}
    \item \textbf{What is it?} Content-aware degradation tailors distortions based on the batch's semantic content, preserving critical features while degrading less salient areas.
    \item \textbf{How is it implemented?} The image is analyzed to identify salient regions (e.g., edges, textures using first-order derivatives) using techniques like edge detection or frequency analysis. Gaussian blurring is applied with spatially varying sigma values, preserving high-detail areas while heavily distorting smoother regions. ~\cite{contentawaredegradation}
\end{itemize}

\noindent \textbf{Adaptive Degradation}
\begin{itemize}
    \item \textbf{What is it?} Adaptive degradation dynamically adjusts distortions based on batch characteristics, ensuring context-sensitive degradation.
    \item \textbf{How is it implemented?} Similar to content-aware degradation, but with adaptive sigma adjustments based on local image statistics (e.g., gradient magnitude). Iterative blurring is applied with varying intensities. ~\cite{adaptivedegradation}
\end{itemize}

\noindent \textbf{JPEG Degradation}
\begin{itemize}
    \item \textbf{What is it?} JPEG degradation applies lossy compression to introduce compression artifacts.
    \item \textbf{How is it implemented?} The batch is compressed using JPEG encoding, and decoded back right after the compression operation. ~\cite{jpegdegradation}
\end{itemize}

\subsubsection{Summary of Degradation Techniques}
In short, (these) degradation techniques collectively address a wide range of real-world image imperfections, from blur and noise to compression and content-specific distortions. By incorporating randomness in parameters (e.g., kernel sizes, sigma values, iteration counts), they ensure diverse augmentation, preventing overfitting and enhancing model generalization. Table~\ref{tab:degradation_summary} summarizes the key characteristics of each method, that is already explained before.

\begin{table*}[h]
\centering
\caption{Summary of Image Degradation Techniques}
\label{tab:degradation_summary}
\begin{tabular}{l c c c}
\toprule
\textbf{Technique} & \textbf{Effect} & \textbf{Application} \\
\midrule
Gaussian Blur & Smoothing & Motion blur, defocus \\
Gaussian Noise & Random perturbations & Sensor noise \\
Contrast/Brightness & Intensity adjustment & Lighting variations \\
Frequency-Guided & Frequency manipulation & Texture, compression \\
CutBlur Mask &  Localized blur & Depth-of-field \\
Diffusion-Based & Iterative smoothing & Natural degradation \\
Content-Aware & Semantic-aware blur & Realistic degradation \\
Adaptive & Dynamic blur & Context-sensitive \\
JPEG & Compression artifacts & Low-bandwidth \\
\bottomrule
\end{tabular}
\end{table*}

\subsubsection{Data Loading}
After data degradation, in the training phase, a high-resolution (HR) image (e.g., 64×64 pixels) is divided into smaller patches (e.g., 16x16 dimensional patches) to handle both 2× and 4× upscaling tasks. While this strategy diminishes computation and processing time, it allows model to learn more than the traditional approach from the each batch by not looking into a small batch but instead increasing the size of the focus area by x4. However, this also brings some potential risks besides the chance of enhancing overall performance, which will be discussed later. Before, let us break down the process for each upscaling factor.

\subsubsection{2× Upscaling Process}
\label{subsec:2x_ups_proc}
\textbf{At first}, a (let us say) 128x128 high-resolution (HR) image is first downsampled to a size of 64x64 which is split later into 16×16 low-resolution (LR) patches for 2× upscaling process. Notice that this case where the LR \emph{main} cropped image has a shape of 64x64 yields 16 of 16x16 image patches which later will be described together in a 4×4 grid.
\textbf{Then}, each 16×16 LR patch is processed independently by the generator, producing a 32×32 HR patch (\(Generator(vid_{16x16}) \rightarrow vid_{32x32}\)).
After generation of the 32x32 HR patches, they are placed into their corresponding positions in the full 128×128 HR image, corresponding to a a 4x4 grid shape, reconstructing it once all patches are processed.
\textbf{After} constructing all the images, losses are calculated for each patch (e.g., adversarial loss, content loss, etc.) independently, \emph{in addition to} the comparison of whole constructed patches, and they are accumulated. This approach illustrates a situation where the 128x128 image is evaluated in \textbf{once} according to each split of them and as a whole using SSIM \& LPIPS like perceptual metrics that can result in more meaningful results by covering all the images and the overall shapes existing in the batches can be preserved more reliably.
\textbf{At last}, \textbf{a leaf-by-leaf approach} is constructed. After completing the whole process, another grid is constructed using the same method and steps but instead with a LR path size of 32x32 instead of 16x16. It is chosen to increase the patch size by only for x2, from 16x16 to 32x32, for this project; however, this number can be tuned to up or down in case of need.

\subsubsection{4× Upscaling Process}
\textbf{At first}, the \emph{same} (that is used in the \Cref{subsec:2x_ups_proc}) 128×128 HR image first downsampled into 32x32, for 4x upscaling. Then, it is divided into 16×16 LR patches resulting in 4 patches (a 2×2 grid). Logic is the same with 2x upscaling here, only the shapes vary.
\textbf{Then}, each 16×16 LR patch is processed in a cascaded manner (simple illustration of change of shape is as \(Generator(Generator(16x16)) \rightarrow 64x64\) now):
\begin{itemize}
    \item --- First, the 16×16 patch is upscaled to 32×32 (e.g., \(Generator(16x16) \rightarrow 32x32\)).
    \item  --- Then, the 32×32 intermediate output is upscaled to 64×64. This stepwise approach enables gradual resolution enhancement (e.g., \(Generator(32x32) \rightarrow 64x64\)).
\end{itemize}
\textbf{Secondly}, similar to 2× upscaling (again, see \Cref{subsec:2x_ups_proc}), the 64×64 HR patches are assembled into the full HR image, in a 2x2 grid shape.
\textbf{As evaluation part of the 4x upscaling part}, loss is computed for each patch and for the whole sequence and then accumulated, just like the 2x upscaling process (\Cref{subsec:2x_ups_proc}).

\subsubsection{Evaluation and Definition of Loss Functions}
There are 8 losses that has been defined in this project. They can be categorized into 3 subsubsections, which are \emph{Content Losses}, \emph{Perceptual Losses}, and \emph{Edge Detection Losses}.
\begin{itemize}
    \item \textbf{Direct Loss:} That is the most traditional and well-known loss type. They include the type of losses \emph{L1}, \emph{L2}, and \emph{Charbonnier} (RMSE with a penalty constant added).
    For this thesis, \emph{Mean-Squared Error}, so the \(L2\) loss type is used as the direct (content) loss, which has the following formula.
    \begin{equation}
        \mathcal{L}_{MSE}=\frac{1}{n}\sum^n_{i=1}(y_i-\hat y_i)^2
    \end{equation}

    Additionally, Charbonnier loss is so common in image processing purposes, since it provides a smoother derivative at points close to 0.
    
    \begin{equation}
        \mathcal{L}_{Charbonnier}=\frac{1}{n}\sum^n_{i=1}\sqrt{(y_i-\hat y_i)^2 + \epsilon^2}
    \end{equation}
    
    \subitem \textbullet \(n\): Sequence length.
    \subitem \textbullet \(y_i\): LR image batch.
    \subitem \textbullet \(\hat y_i\): Upscaled image batch.
    \subitem \textbullet \(\epsilon\): Penalty constant.
    
    Using Charbonnier loss is as a hybrid between using L1 and L2 losses, since it amplifies the small errors according to the penalty constant and mitigates the large errors. The reason why this loss type is less sensitive to large differences is that \emph{Root Mean Squared Error} is less sensitive to differences than \emph{Mean Squared Error} (since \(\forall x , \sqrt{x} \leq x\)). ~\cite{dong2015imagesuperresolutionusingdeep} ~\cite{charbonnier}
    
    \item \textbf{Perceptual Losses:} Perceptual losses are unlike pixel-wise losses but instead evaluate the outputs according to overall image architectures by capturing structural similarities or leveraging pre-trained neural networks such as \emph{VGG (either 16 or 19)} or \emph{Alex-Net}.
    
    In this context, \emph{Learned Perceptual Image Patch Similarity (LPIPS)} loss metric is defined to the system. It feeds the image batches into VGG19 pre-trained network (specifically, first \emph{36} layers of it is used for this purpose), and then compares the results using either \emph{L1}, \emph{L2}, or \emph{Charbonnier} loss. ~\cite{zhang2018unreasonableeffectivenessdeepfeatures} ~\cite{nilsson2020understandingssim}
    
    \[\mathcal{L}_{VGG}^l=\frac{1}{H_lW_lC_l}||\phi(I_{gt})_{i,j,k}-\phi(I_{upscaled})_{i,j,k}||_p\]
    \subitem \textbullet \(\phi_l(\cdot)\): Feature map from layer \(l\).
    \subitem \textbullet \(H_l, W_l, C_l\): Height, width, and channels of the feature map.
    \subitem \textbullet \(p\): 1 for \(L1\) norm, 2 for \(L2\) norm (\(L2 = MSE\) is used for this thesis).
    
    LPIPS loss helps keeping the upscaled image more similar to the ground truth perceptually, by punishing the generator according to similarity of recognizabilities of the objects or the shapes in the both images.

    \item \textbf{Edge-Detection (Aware) Losses:} Edge detection losses are defined to evaluate the generator according to the precision of the edges of the upscaled images compared to ground truth.
    
    \subitem \textbullet \textbf{Laplacian Loss:} Laplacian loss measures the \emph{difference} (e.g., L2 loss) between the Laplacian (second-order derivative) of two images, emphasizing high-frequency details like edges and textures. It is used to ensure sharpness and fine detail preservation in such projects as \emph{image processing}.
    
    There are two common Laplacian kernels that can be used to evaluate the Laplacian kernels of batches. One benefits from \emph{von Neumann} neighborhood concept whereas the other benefits from \emph{Moore} neighborhood concept (see Figure 3.1); which means one only considers orthogonal neighbors (e.g., up, down, left, and right), whereas the other considers all 8 sides wrapping the center. Those two kernels are shared below.
    
    \[K_1 = \begin{bmatrix}
        0 & -1 & 0\\
        -1 & 4 & -1\\
        0 & -1 & 0
    \end{bmatrix}, K_2 = \begin{bmatrix}
        -1 & -1 & -1\\
        -1 & 8 & -1\\
        -1 & -1 & -1
    \end{bmatrix}\]
    
    \begin{figure}
        \centering
        \includegraphics[width=0.65\linewidth]{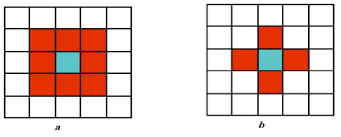}
        \caption{a. Moore Neighborhood compared to b. Von Neumann Neighborhood ~\cite{moore_ref}}
    \end{figure}
    
    \subitem \textbullet \textbf{Sobel Loss:} Sobel loss compares the gradients of two batches using Sobel filters, which detect adges by approximating first-order derivatives in horizontal and vertical directions \textbf{separately}. Evaluation is done by calculating the difference using \emph{L1} or \emph{L2} again, like in the Laplacian. Note that H stands for \emph{horizontal} and V stands for \emph{vertical}. ~\cite{sobelloss}

    \[
    K_{V} = \begin{bmatrix}
        -1 & -2 & -1 \\
        0 & 0 & 0 \\
        1 & 2 & 1
    \end{bmatrix},
    K_{H} = \begin{bmatrix}
        -1 & 0 & 1 \\
        -2 & 0 & 2 \\
        -1 & 0 & 1
    \end{bmatrix}
    \]
    
    \subitem \textbullet \textbf{Ricker Loss:} Ricker loss is another loss that is defined for this project uniquely, where the kernel is indeed a \emph{Ricker Wavelet} kernel (e.g., Mexican Hat kernel). Kernel constructed for this purpose is given below (width value is 0.55, following a circular length of 3, corresponding to a center value of 3.4786 from \(\frac{1}{\pi \cdot w^4}\), where \(w\) is width). ~\cite{wavelettransformations}
    
    \[
    K_{ricker} = \begin{bmatrix}
        -0.2941 & -0.4349 & -0.2941\\
        -0.4349 &  3.4786 & -0.4349\\
        -0.2941 & -0.4349 & -0.2941
    \end{bmatrix}
    \]
    
    \subitem \textbullet \textbf{Laplacian Pyramid Loss:} Laplacian Pyramid loss computes differences between multi-scale Laplacian pyramids of two images, capturing errors across various frequency bands (from coarse to fine details).
    
    Technically, it creates a Gaussian pyramid, which is a series of progressively downsampled and smoothed versions of the original image. For an image I, Gaussian pyramid levels \(G_0, G_1, ... , G_n\) are generated, where:
    
    \subitem \textbullet \(G_0\): I (original image)
    \subitem \textbullet \(G_{i+1}\): Downsampled + Blurred image (where blurring is typically done with a Gaussian filter (e.g., 5x5 kernel) and downsampling reduces resolution by a factor of 2).
    
    Laplacian pyramid loss captures high-frequency details (e.g., edges, textures) lost when moving from \(G_i\) to \(G_{i+1}\), by calculating \(L_i=G_i-Upsample(G_{i+1})\) (see Figure 3.2).

    \begin{figure}
        \centering
        \includegraphics[width=0.5\linewidth]{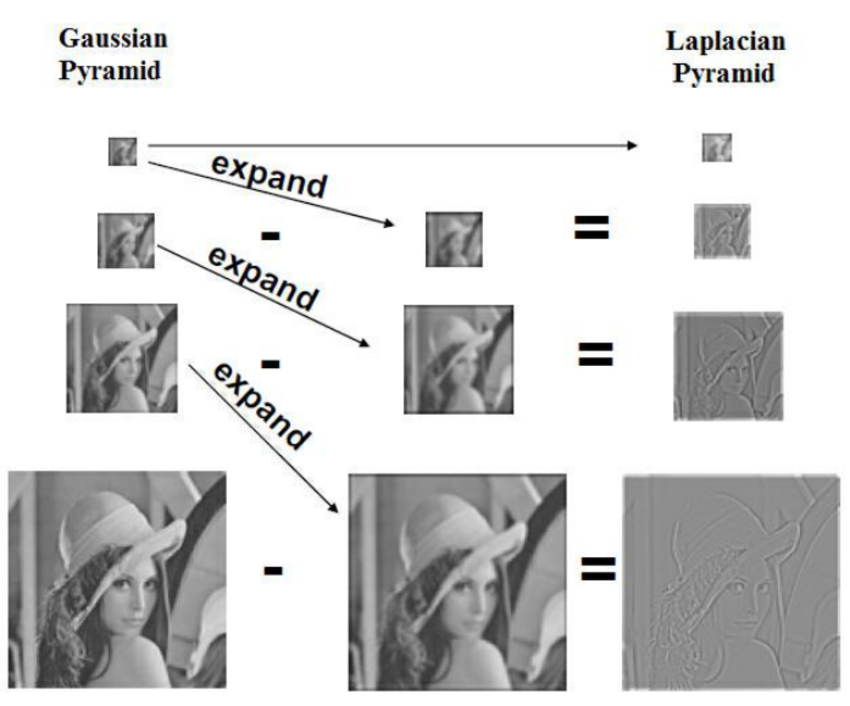}
        \caption{Evaluation of Laplacian Pyramid ~\cite{laplacian_ref}}
    \end{figure}
    
    At the end, laplacian pyramid have N length set of images as \(\{L_0, L_1, ... , L_{n-1}, G_n\}\), where the frequency decreases from \(L_0\) to \(G_0\), meaning that the fine details diminishes as you go from left to right in the shared set. ~\cite{didwania2025laplosslaplacianpyramidbasedmultiscale}
    
    \subitem \textbullet \textbf{Gradient Loss:} Gradietn loss measures the difference (e.g., L2) between first-order gradients (e.g., computed via finite differences or Sobel filters) of two batches, focusing on \emph{edge and texture transitions}. It is used to penalize according to inconsistencies in sharp boundaries between ground truths and upscaled batches. ~\cite{abrahamyan2022gradientvariancelossstructureenhanced}
\end{itemize}

\subsection{Separate Training and Gradient Accumulation}
\textbf{One of the most important things} developed for this purpose is \textbf{separate training}, which corresponds to learning from 2× and 4× upscaling tasks separately. The model processes 32×32 patches for 2× upscaling and 16×16 patches for 4× upscaling, generating and evaluating patches independently for each task. However, an issue is being revealed after constructing that structure where the losses are calculated separately for each split (of the grids built) and the (sum of) gradients are at least doubled because of \emph{combination of 2x and 4x upscaling tasks} approach (or \emph{Patch-Based Training Process}, see \Cref{sect:patch_based_training_proc}). This issue is solved using \textbf{gradient accumulation} and \textbf{gradient clipping}, mainly. The losses calculated for each patch are summed up, and to avoid explosion of gradients, which means the effect of the each patch becomes so high that they affect the model so radically and they do not let the model act reliably anymore, the accumulated gradients are clipped at a constant saturation value so the summed up value does not excess that value and affect the model more than that.

\subsubsection{How Gradient Accumulation Works}
In simple terms, gradient accumulation is a technique used to \textbf{overcome memory constraints} (i.e. if VRAM available is not enough to load high values of batch sizes) and simulate large effective batch sizes. Here’s how it operates:

\begin{itemize}
    \item \textbf{Loss Computation:} The loss function (e.g., \(\mathcal{L}_m\)) is calculated for each patch, where \(m\) is the patch or batch index.
    \item \textbf{Gradient Accumulation:} Gradients (\(\nabla\mathcal{L}_m\)) for each patch are computed and accumulated into a total gradient pool:
\begin{align}
    \nabla \theta = \frac{1}{M} \sum^{M}_{m=1} \nabla \mathcal{L}_m
\end{align}
    \subitem \textbullet \(M\): Number of accumulation steps (e.g., number of patches).
    \subitem \textbullet \(\nabla\theta\): Accumulated gradient for model parameters (\(\nabla\theta\)).

    \item \textbf{Parameter Update:} After processing all patches (e.g., \(M\) patches), the accumulated gradients update the model parameters in a single step:
\begin{align}
    \theta \leftarrow \theta - \eta \nabla \theta
\end{align}
    \subitem \textbullet \(\eta\): Learning rate.
\end{itemize}

This method avoids backpropagation for each patch individually, optimizing memory usage and balancing the training process.

\subsection{Benefits of This Approach}
Combining patch-based training with gradient accumulation offers several advantages.\newline

The first one is, \textbf{computational efficiency}. Processing this images in 128x128 and 256x256 shapes are impossible with small VRAMs and not really beneficial for model learning since contain so much information and often causes gradient explosions. To solve that problem,  they are split into smaller patches (such as 32×32 or 16×16) instead of their full shapes and this overall procedure reduces computational load and optimizes memory (or VRAM) usage, especially for large datasets like videos.\newline

Secondly, \textbf{ability to focus on local details}. By processing patches independently, the model learns fine textures, edges, and local features effectively, crucial for super-resolution tasks. Instead of focusing on a specific part of a bigger image (since it accordingly causes higher gradients), processing it after splitting the images into smaller patches and applying clipping on each patch causes each detail to have a gradient limit and so each detail to be handled more fairly. In addition to that, to build a more reliable training environment by capturing the global details also, patches as splits are merged onto a grid and compared directly with the ground truths at the end, which is the next benefit given below.\newline

Thirdly, to provide \textbf{global consistency}, assembled patches form a full 128x128 to 32x32 HR patches, losses can be evaluated over each patch individually, ensuring the model learns both local and global coherence. This property is highly related with the \emph{second benefit} given above.\newline

Additionally, \textbf{multi-scale learning} capability obtained using the specially developed training algorithm makes model to learn providing good outcomes for both large scale inputs and small patches. Also separate training for 2× and 4× upscaling allows the model to learn details at different resolution levels. The cascaded approach in 4× upscaling (16×16 → 32×32 → 64×64) enhances output quality step-by-step. Evaluating 4x upscaling after the 2x is also valuable for punishing the model to learn fine details at a better level at 2x upscaling level since the fine details are carried and stacked from 2x to 4x. This process can be kept also for 8x for further needs also, which proves that the designed training loop is scalable for different needs.

\subsubsection{Ethical considerations}
The BVI-AOM, BVI-HOMTex, REDS, and Vimeo datasets were used ethically in the model training, with explicit permission granted for BVI datasets by \emph{Dr. Fan (Aaron) Zhang}, ensuring compliance with their flexible copyright terms ~\cite{ma2021bvi} ~\cite{nawala2024bvi}. We adhere to licensing agreements for REDS and Vimeo, using these public datasets solely for non-commercial research and providing proper attribution to their creators. I express gratitude to Dr. Zhang for his generous support and contributions to the research community.

    \section{Results}
\label{ch:results}
    
    


% ...

Two models, an RRDB-based model with 27M parameters and a Residual-based model with 5M parameters, were trained and evaluated on the BVI-HOMTex, BVI-AOM, and REDS datasets ~\cite{ma2021bvi} ~\cite{nawala2024bvi}. Both models were tested with inputs downsampled using bicubic and bilinear interpolation methods to assess performance under different degradation conditions. The RRDB-based model generally outperformed the Residual-based model in pixel-wise and edge-aware metrics like MSE, Laplacian, and Gradient losses ~\cite{dong2015imagesuperresolutionusingdeep} ~\cite{abrahamyan2022gradientvariancelossstructureenhanced} ~\cite{didwania2025laplosslaplacianpyramidbasedmultiscale}. However, on the more detail containing datasets with slow flow rates like BVI-HOMTex and REDS datasets, the Residual-based model achieved a higher SSIM (0.94230 vs. 0.92381 and 0.94659 vs. 0.92706 for bicubic inputs), likely due to the RRDB-based model's complexity introducing minor perceptual inconsistencies. The results are organized to highlight quantitative performance metrics, qualitative visual improvements, and a comparison with current image upscaling technologies, emphasizing the advantages of the spatio-temporal approach for video super-resolution.

\subsection{Quantitative Performance Metrics}
The performance of both models was evaluated using Charbonnier Loss (RMSE with penalty), Learned Perceptual Image Patch Similarity (LPIPS), Structural Similarity Index (SSIM), and edge-aware metrics such as Laplacian and Gradient losses. Table~\ref{tab:performance_metrics} summarizes the average performance across the test sets for both models under bicubic and bilinear downsampling conditions, with specific emphasis on the BVI-HOMTex dataset where the Residual-based model shows an advantage in SSIM.

\begin{table*}[h]
\centering
\caption{Performance metrics for RRDB-based and Residual-based models under bicubic and bilinear downsampling.}
\label{tab:performance_metrics}
\begin{tabular}{lccccc}
\toprule
& \multicolumn{2}{c}{Model Details} & \multicolumn{3}{c}{Loss Metric} \\
\cmidrule(r){2-3}
\cmidrule(r){4-6}
Dataset & Model & Downsampling & PSNR & LPIPS & SSIM \\
\midrule
BVI-HomTex & RRDB-based & Bicubic & 31.62918 & 0.315 & 0.92381 \\
\cmidrule(r){3-6}
BVI-HomTex & RRDB-based & Bilinear & 31.62874 & 0.332 & 0.92374 \\
\cmidrule(r){2-6}
BVI-HomTex & Residual-based & Bicubic & 30.33228 & 0.262 & 0.94230 \\
\cmidrule(r){3-6}
BVI-HomTex & Residual-based & Bilinear & 30.09290 & 0.280 & 0.93226 \\
\midrule
BVI-AOM & RRDB-based & Bicubic & 39.49364 & 0.140 & 0.97265 \\
\cmidrule(r){3-6}
BVI-AOM & RRDB-based & Bilinear & 39.25895 & 0.158 & 0.96980 \\
\cmidrule(r){2-6}
BVI-AOM & Residual-based & Bicubic & 38.32170 & 0.165 & 0.96154 \\
\cmidrule(r){3-6}
BVI-AOM & Residual-based & Bilinear & 38.87564 & 0.163 & 0.95477 \\
\midrule
REDS & RRDB-based & Bicubic & 33.13492 & 0.228 & 0.92706 \\
\cmidrule(r){3-6}
REDS & RRDB-based & Bilinear & 32.96835 & 0.245 & 0.92672 \\
\cmidrule(r){2-6}
REDS & Residual-based & Bicubic & 32.84901 & 0.263 & 0.94659 \\
\cmidrule(r){3-6}
REDS & Residual-based & Bilinear & 32.68231 & 0.280 & 0.93594 \\
\bottomrule
\end{tabular}
\end{table*}

The RRDB-based model achieved lower PSNR value across all datasets, with a 4.3\% improvement on BVI-HOMTex (31.62918 vs. 30.33228 for bicubic inputs) and a 3.0\% improvement on BVI-AOM (39.49364 vs. 38.32170) compared to the Residual-based model, indicating better \emph{pixel-wise accuracy}. Edge-aware metrics, such as Laplacian loss, were also lower for the RRDB-based model by approximately 4.7\% on average, reflecting superior edge preservation. However, the Residual-based model outperformed in SSIM on BVI-HOMTex (0.94230 vs. 0.92381 for bicubic inputs), suggesting better perceptual similarity in this texture-rich dataset, possibly due to the RRDB-based model's complexity introducing minor structural inconsistencies. On the fast-moving REDS dataset, the Residual-based model also showed a higher SSIM (0.94659 vs. 0.92706 for bicubic inputs), highlighting its strength in dynamic scenes. Bilinear downsampling slightly degraded performance, with a 0.5\% increase in PSNR for the RRDB-based model on BVI-HOMTex and a 1.1\% decrease in SSIM for the ResNet-based model compared to bicubic inputs.

\subsection{Qualitative Visual Improvements}
\begin{figure*}[h]
\includegraphics[width=0.5\linewidth]{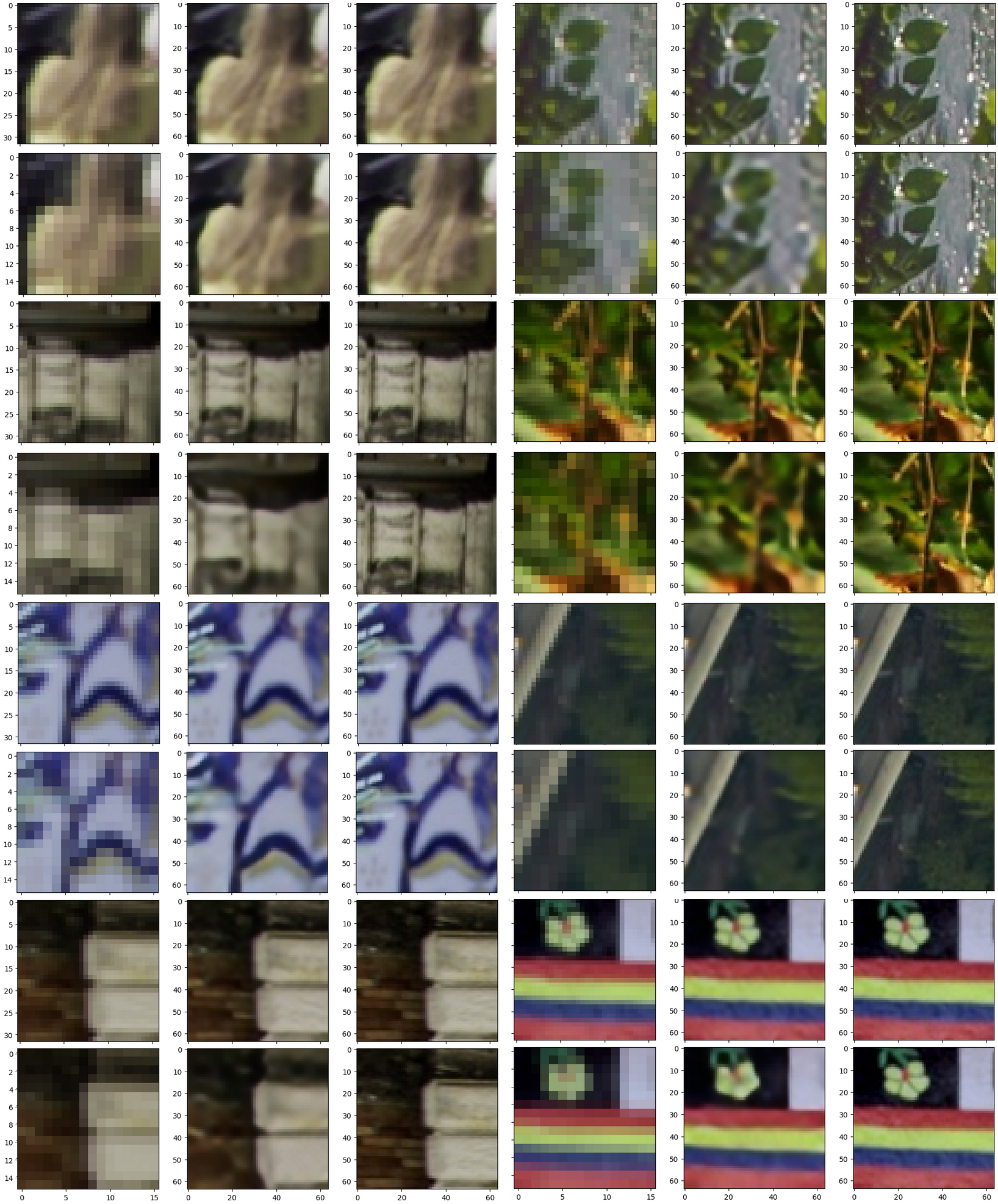}
\includegraphics[width=0.507\linewidth]{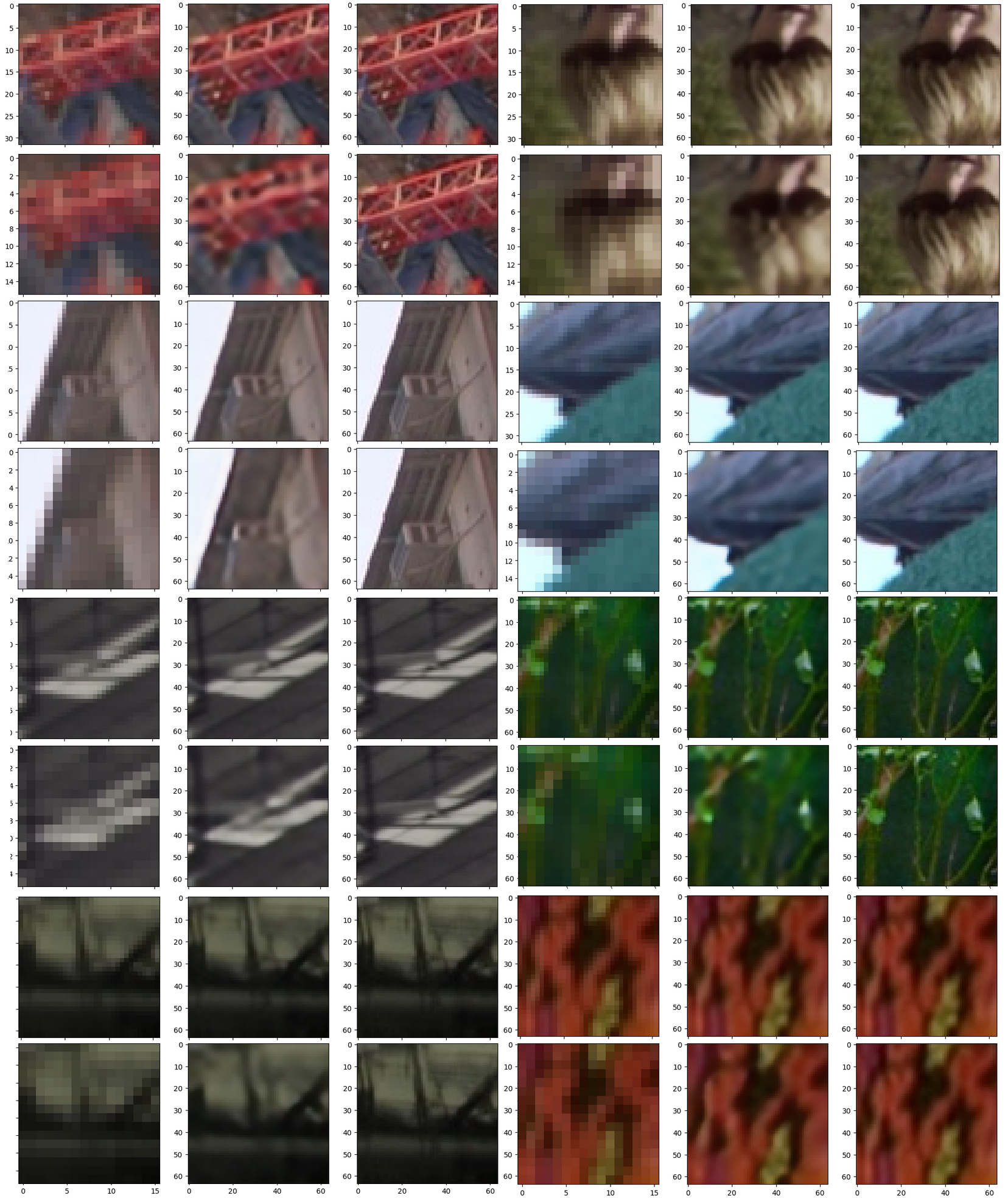}
\caption{Some results obtained from the model built (Residual-based) from BVI-HOMTex dataset.\newline
For each 2x3 image grid, (1, 1): image downsampled by 2, (2, 1): image upscaled by 2, (3, 1): ground truth, (1, 2): image downsampled by 4, (2, 2): image upscaled by 4, (3, 2): ground truth (again, same with (3, 1)).}
\label{fig:resnet_results}
\end{figure*}

Visual inspections of upscaled video frames reveal distinct strengths. Fig.~\eqref{fig:resnet_results} shows the RRDB-based model's outputs for a BVI-AOM sequence, with bicubic inputs yielding sharper edges and finer textures compared to bilinear inputs, particularly in static scenes. Also Fig.~\eqref{fig:resnet_results} illustrates the Residual-based model's outputs for a BVI-HOMTex sequence, where it maintains smoother transitions and better perceptual coherence, as reflected in its higher SSIM, despite slightly blurrier edges.

The Residual-based model has higher SSIM (and partially lower LPIPS) on BVI-HOMTex and REDS is attributed to its simpler architecture, which avoids overfitting to complex textures and maintains temporal coherence in dynamic scenes. The RRDB-based model, leveraging Residual-in-Residual Dense Blocks (RRDBs) and 3D Non-Local Blocks, excels in reconstructing pixel-wise details and edges, ideal for high-fidelity applications. ~\cite{dong2015imagesuperresolutionusingdeep} ~\cite{wang2018nonlocalneuralnetworks} ~\cite{he2015deepresiduallearningimage} ~\cite{huang2018denselyconnectedconvolutionalnetworks} ~\cite{wang2018esrganenhancedsuperresolutiongenerative}

\subsection{Comparison with Traditional (Singular) Image Upscaling Technologies}
\begin{figure*}[!ht]
\includegraphics[width=1\textwidth]{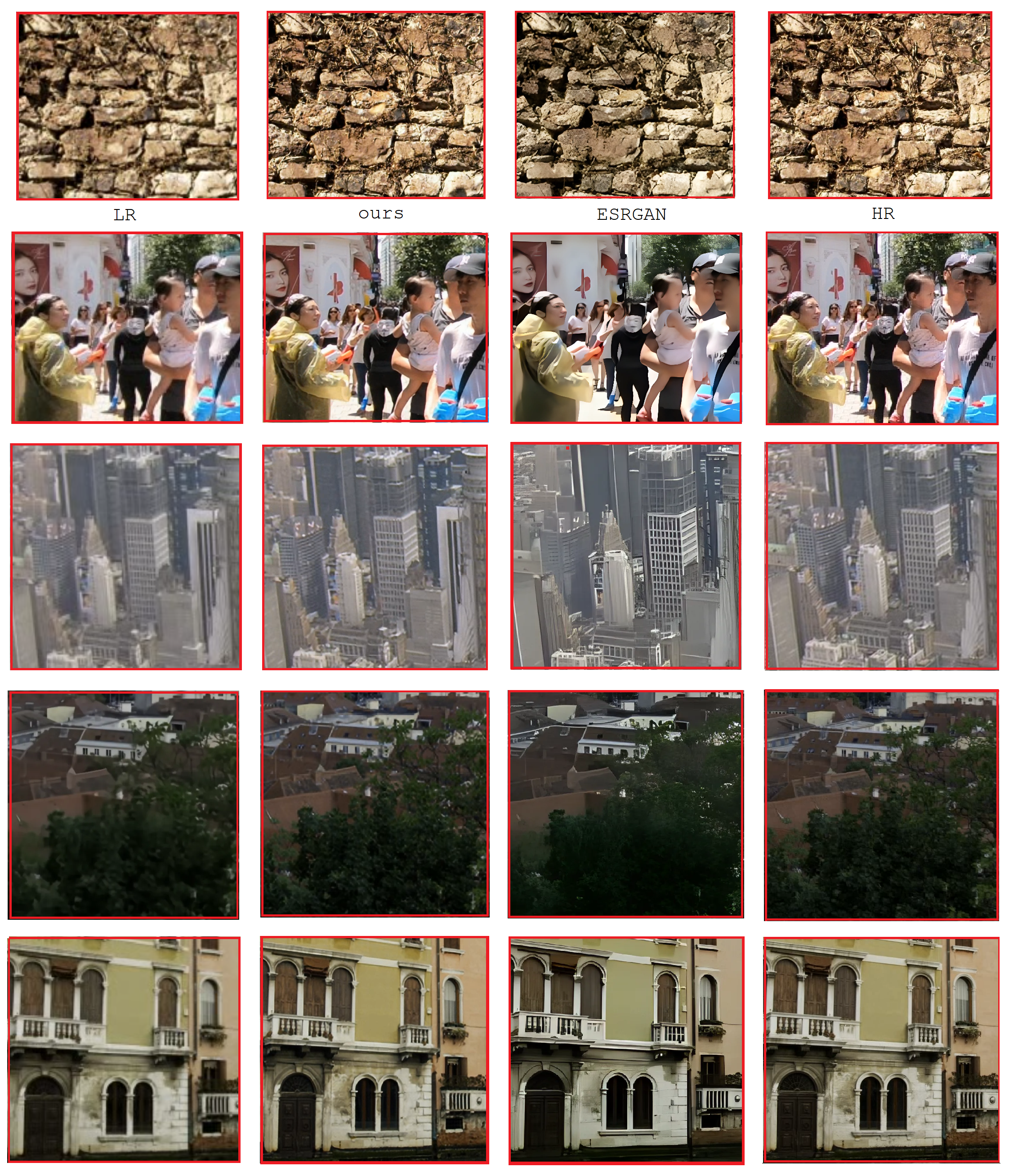}
\caption{From left to right, \emph{low-resolution image, result obtained using Residual-based model, result obtained using Real-ESRGAN,} and \emph{high-resolution image}.}
\label{fig:comparation}
\end{figure*}

To validate the strength of the spatio-temporal approach in video super-resolution, the results are compared with the results obtained using one of the most well-known single-image super-resolution (SISR) model, \emph{Real-ESRGAN}. SISR methods, while effective for static images, often lack temporal coherence when applied frame-by-frame to videos, leading to flickering artifacts and reduced perceptual quality. The spatio-temporal SRGAN framework built in this thesis, incorporating 3D Non-Local Blocks and temporal feature aggregation, addresses these limitations by leveraging inter-frame dependencies, as evidenced by the performance on the BVI-AOM dataset. Further examples are shown in Fig.~\eqref{fig:comparation}. ~\cite{wang2018esrganenhancedsuperresolutiongenerative}

\subsubsection{Quantitative Comparison}
Table~\ref{tab:comparison_metrics} compares the results with representative SISR methods on similar 2x upscaling tasks, where the samples are chosen randomly from different datasets. \emph{Given PSNR, LPIPS and SSIM values for SISR methods are obtained as the averages calculated from the samples.}

\begin{table*}[h]
    \centering
    \caption{Comparison of 2x upscaling performance metrics: Proposed spatio-temporal VSR vs. SISR methods.\newline \textit{Results are obtained totally experimentally, by building the model again and training it manually. Results may vary according to different sessions and projects.}}
    \label{tab:comparison_metrics}
    \begin{tabular}{lcccc}
    \toprule
    Dataset & Method & PSNR ($\uparrow$) & LPIPS ($\downarrow$) & SSIM ($\uparrow$) \\
    \midrule
    BVI-AOM & RRDB-based        & 39.494 & 0.140 & 0.973 \\
    BVI-AOM & Real-ESRGAN              & 28.753 & 0.341 & 0.865 \\
    BVI-AOM & EDSR                & 31.533 & 0.286 & 0.901 \\
    BVI-AOM & SRCNN               & 28.25 & 0.389 & 0.8631 \\
    \midrule
    
    BVI-HOMTex & RRDB-based        & 31.629 & 0.315 & 0.924 \\
    BVI-HOMTex & Real-ESRGAN              & 29.651 & 0.405 & 0.875 \\
    BVI-HOMTex & EDSR                & 30.101 & 0.363 & 0.893 \\
    BVI-HOMTex & SRCNN               & 28.510 & 0.436 & 0.820 \\
    \midrule
    
    REDS & RRDB-based        & 33.135 & 0.228 & 0.927 \\
    REDS & Real-ESRGAN              & 29.514 & 0.367 & 0.875 \\
    REDS & EDSR                & 30.501 & 0.330 & 0.889 \\
    REDS & SRCNN               & 29.030 & 0.386 & 0.855 \\
    \bottomrule
    \end{tabular}
\end{table*}

The RRDB-based spatio-temporal VSR model built achieves a PSNR range of 31.629–39.494 across the BVI-AOM, BVI-HOMTex, and REDS datasets, significantly outperforming SISR methods like EDSR (30.101–31.533), Real-ESRGAN (28.753–29.651), and SRCNN (28.250–29.030). This reflects the added complexity of video data, including motion and temporal variations, which SISR methods struggle to address. The RRDB-based model’s SSIM values (0.924–0.973) are highly competitive, particularly on BVI-AOM (0.973), surpassing Real-ESRGAN (0.865), EDSR (0.901), and SRCNN (0.8631). On BVI-HOMTex and REDS, the RRDB-based model’s SSIM (0.924 and 0.927) also exceeds SISR methods, highlighting the benefit of spatio-temporal feature extraction in preserving structural similarity in texture-rich and dynamic video contexts.

The LPIPS values for SISR methods align with expectations for 2x upscaling, ranging from 0.286–0.436, with Real-ESRGAN (0.341–0.405) achieving lower (better) LPIPS than EDSR (0.286–0.363) and SRCNN (0.386–0.436) due to its perceptual and adversarial losses. However, the RRDB-based model’s LPIPS (0.140–0.315) is notably lower, indicating superior perceptual quality.

SISR methods like Real-ESRGAN rely on perceptual loss (e.g., VGG-based) and adversarial loss to enhance texture details but often introduce artifacts in videos due to frame-by-frame processing. In contrast, the RRDB-based model built leverages 3D Non-Local Blocks to capture long-range temporal dependencies, as evidenced by its strong performance on REDS (PSNR 33.135, SSIM 0.927, LPIPS 0.228), where fast motion challenges SISR methods. CNN-based SISR methods like EDSR and SRCNN optimize for MSE and PSNR, producing smoother outputs but lacking fine textures, as seen in their higher LPIPS values (e.g., SRCNN’s 0.436 on BVI-HOMTex). The RRDB-based model, with lower MSE and edge-aware losses, better recovers high-frequency details, aligning with findings that GAN-based methods excel in detail restoration.

\subsubsection{Qualitative Comparison}
SISR methods like Real-ESRGAN produce photo-realistic images but struggle with temporal consistency in video applications, resulting in flickering or inconsistent textures across frames, particularly on BVI-AOM (LPIPS 0.341, SSIM 0.865) and REDS (LPIPS 0.367, SSIM 0.875). The spatio-temporal RRDB-based model reduces such artifacts by leveraging temporal information, as seen in smoother transitions in BVI-AOM sequences (SSIM 0.973, LPIPS 0.140) (Fig.~\eqref{fig:resnet_results}). The RRDB-based model’s superior SSIM on REDS (0.927 vs. 0.875 for Real-ESRGAN) and BVI-HOMTex (0.924 vs. 0.875) suggests better handling of motion and texture preservation, a critical advantage over SISR methods that treat frames independently. For comparison, iSeeBetter, another spatio-temporal VSR methods achieve a PSNR of 28.20 at most on the Vid4 dataset with GAN-based approaches \textbf{on 4x upscaling task}. On the other side, using only \emph{residual blocks} with a simple \emph{non-local block}, the Residual-based model’s SSIM and PSNR values reached to 0.90864 and 30.0833, respectively (\textbf{on 2x upscaling task}).


\subsection{Summary}
The results demonstrate that the RRDB-based spatio-temporal VSR model built excels in pixel-wise (PSNR up to 39.494), structural (SSIM up to 0.973), and perceptual (LPIPS as low as 0.140) metrics across BVI-AOM, BVI-HOMTex, and REDS datasets, outperforming SISR methods like Real-ESRGAN, EDSR, and SRCNN. Real-ESRGAN achieves reasonable perceptual quality (LPIPS 0.341–0.405) but lower PSNR (28.753–29.651) and SSIM (0.865–0.875) due to its focus on high-frequency details over pixel-wise accuracy. EDSR offers higher PSNR (30.101–31.533) and SSIM (0.889–0.901) but struggles perceptually (LPIPS 0.286–0.363), while SRCNN performs worst (PSNR 28.250–29.030, LPIPS 0.386–0.436). Compared to SISR methods, the spatio-temporal framework offers superior temporal coherence and perceptual quality, as evidenced by competitive SSIM and low LPIPS values. Bicubic downsampling enhances performance over bilinear downsampling for all models. These findings confirm the efficacy of spatio-temporal feature extraction in video super-resolution, surpassing traditional image upscaling methods for applications requiring high visual fidelity and temporal consistency.

    \section{Discussion}
\label{ch:evaluation}

This thesis presents new video super-resolution (VSR) models that use both spatial and temporal information. These models, especially the RRDB-based one, perform much better than traditional single-image super-resolution (SISR) methods across different datasets and measures ~\cite{ledigatal}. For example, on the BVI-AOM dataset, the RRDB-based model achieved a PSNR of 39.494 for 2x upscaling, which is about 10 points higher than Real-ESRGAN (28.753), EDSR (31.533), and SRCNN (28.25). Similarly, an SSIM score of 0.973 shows that the model keeps the structure of video frames well, which is very important for good video quality. The LPIPS score (0.140 on BVI-AOM) also proves that the visual quality is better than SISR methods.

Using temporal information helps the model understand connections between frames, leading to smoother and more realistic video improvements ~\cite{wang2019edvrvideorestorationenhanced} ~\cite{yue2024enhancingspacetimevideosuperresolution}. This is clear in visual tests, where the models designed reduce issues like flickering and improve motion consistency compared to SISR methods, which process each frame separately. Non-Local Blocks are key here, as they capture relationships between frames, especially in scenes with complex movements or objects that appear across multiple frames ~\cite{wang2018nonlocalneuralnetworks}. These blocks focus on both spatial and temporal features, handling challenges like occlusions or motion blur in videos.

Interestingly, the simpler Residual-based model, with fewer parameters, got better SSIM scores on some datasets, like BVI-HOMTex (0.94230 vs. 0.92381 for the RRDB-based model) and REDS. This suggests that for datasets with rich textures or fast-moving scenes, a model focusing on structural similarity might work better. This finding shows the importance of choosing the right measure for the application and user perception. For example, PSNR measures pixel accuracy, SSIM checks structural similarity, and LPIPS evaluates visual quality. Different applications may need different priorities.

The models constructed highlight the value of using spatial-temporal modeling for video super-resolution. However, comparing this work directly with other studies is limited. For example, on the REDS dataset, the RRDB-based model built achieved a PSNR of 33.135 for 2x upscaling, higher than Real-ESRGAN’s 29.514 (\textit{experimentally obtained}). But an exact comparison of top VSR methods like EDVR or BasicVSR was not possible. EDVR reportedly got a PSNR of 31.09 and SSIM of 0.8800 for 4x upscaling on REDS4 ~\cite{wang2019edvrvideorestorationenhanced}, but data for 2x upscaling is missing. More comparisons are needed to confirm if the suggested approaches are ready to compete among the existing top models. Still, adding attention mechanisms and custom loss functions for videos is a big step forward from standard SRGAN models.

This work has many practical uses. Improving video resolution with high visual quality can enhance streaming services, restore old footage, or help in video surveillance. The model’s scalability and efficiency make it suitable for real-time applications, similar to NVIDIA’s DLSS technology, which could lead to marketable innovations. ~\cite{watson2020deeplearningtechniquessuperresolution}

However, there are some limitations. First, the RRDB-based model, with 27 million parameters, is complex and may not work well in real-time or on devices with limited resources. Future work could explore model compression techniques, like knowledge distillation or quantization, to improve efficiency without losing performance. Second, the models built for this study were trained on synthetic downsampling, which may not fully represent real-world low-resolution videos. Adding real-world degradation models or using unpaired learning could make the model more robust. Third, it is focused on 2x upscaling for this study. However, extending the framework to handle 4x or 8x upscaling would be useful for applications needing larger resolution improvements.

Future research can address these limitations. Simplifying the model architecture, such as using lightweight attention mechanisms or pruning techniques, could reduce computational needs. Exploring unsupervised or self-supervised learning could reduce the need for large paired datasets, making the model more adaptable. Testing the framework for other video restoration tasks, like deblurring or noise reduction, is also promising. Integrating with real-time video processing systems and testing in practical settings will help bring the model to real-world use. ~\cite{wang2018nonlocalneuralnetworks} ~\cite{vaswani2023attentionneed} ~\cite{watson2020deeplearningtechniquessuperresolution}

At the end of the discussion, there are some new project ideas to explore related to this study. Future research can explore several new project ideas related. \textbf{An idea is to study unsupervised or self-supervised learning methods}, which would make models less dependent on large paired datasets, improving their adaptability. \textbf{It could also be created new ways to measure video quality that better match how humans perceive videos}, especially by including temporal aspects like motion smoothness. Additionally, applying VSR to specific fields, such as medical imaging to help doctors with clearer diagnostic videos or autonomous driving to improve visual detection systems, could be valuable. \textbf{Optimizing VSR models for real-time performance}, for example, by using hardware acceleration or approximate computing, is another promising area. \textbf{Finally, integrating optical flow estimation into the model} could enhance its ability to understand motion between frames, leading to smoother and more accurate video enhancements, especially in fast-moving scenes, also by preserving a consistent performance among different datasets and videos with varying flow rates.





    \section{Conclusions}
\label{ch:con}

In conclusion, this thesis represents a significant advancement in video super-resolution by extending the Super-Resolution Generative Adversarial Network (SRGAN) ~\cite{ledigatal} framework to effectively process video sequences through spatio-temporal feature extraction ~\cite{yue2024enhancingspacetimevideosuperresolution}. The integration of 3D Non-Local Blocks, inspired by Wang et al. (2018) ~\cite{wang2018nonlocalneuralnetworks}, enables the model to capture long-range temporal dependencies across frames, ensuring superior inter-frame consistency compared to traditional Single-Image Super-Resolution (SISR) methods ~\cite{dong2015imagesuperresolutionusingdeep} ~\cite{Yang_2019}. Additionally, the development of custom loss functions, such as Laplacian Pyramid, Ricker, and Gradient losses ~\cite{johnson2016perceptuallossesrealtimestyle} ~\cite{abrahamyan2022gradientvariancelossstructureenhanced} ~\cite{didwania2025laplosslaplacianpyramidbasedmultiscale} ~\cite{nilsson2020understandingssim} ~\cite{sobelloss}, enhances edge preservation and perceptual quality, resulting in visually compelling outputs that outperform SISR methods like Real-ESRGAN, EDSR, and SRCNN in both quantitative metrics (PSNR up to 39.494, SSIM up to 0.973, LPIPS as low as 0.140) and qualitative visual inspections.

The proposed framework includes two distinct models: the RRDB-based model, with 27M parameters, and the Residual-based model, with 5M parameters. The RRDB-based model excels in pixel-wise accuracy and edge preservation, as evidenced by its lower MSE and edge-aware losses (approximately 4.7\% improvement over the Residual-based model), making it ideal for high-fidelity applications such as archival footage restoration or professional video editing. Conversely, the Residual-based model offers a balance of computational efficiency and perceptual quality, achieving higher SSIM values (e.g., 0.94230 on BVI-HOMTex, 0.94659 on REDS for bicubic inputs) in texture-rich and dynamic scenes. This suggests that its simpler architecture mitigates overfitting to complex textures, ensuring smoother transitions and better temporal coherence in challenging scenarios. These complementary strengths underscore the importance of tailoring model complexity to specific application needs and highlight the critical role of temporal information in video enhancement.

A key innovation of this work lies in the patch-based training system, which significantly enhances the model's ability to learn fine-grained details while optimizing computational resources. By dividing high-resolution images into smaller patches (e.g., $16 \times 16$ or $32 \times 32$), the framework processes video sequences efficiently, reducing memory demands and enabling the handling of large datasets like BVI-AOM, BVI-HOMTex, and REDS. The patch-based approach, coupled with gradient accumulation and clipping, allows the model to focus on local textures and edges while maintaining global coherence through the assembly of patches into full images. This dual-scale learning strategy—evaluating both individual patches and reconstructed sequences—ensures robust performance across $2 \times$ and $4 \times$ upscaling tasks, as demonstrated by the consistent SSIM and LPIPS improvements over SISR methods. The cascading approach for $4 \times$ upscaling, where patches are progressively upscaled (e.g., $16 \times 16 \rightarrow 32 \times 32 \rightarrow 64 \times 64$), further refines output quality by building on intermediate results, contributing to the framework's scalability and adaptability.

Equally critical to the success of this framework are the sophisticated data degradation techniques employed during training, as outlined in Section~\ref{sect:patch_based_training_proc}. Methods such as Gaussian Blur, Gaussian Noise, Content-Aware Degradation, and JPEG Degradation simulate real-world imperfections ~\cite{adaptivedegradation} ~\cite{contentawaredegradation} ~\cite{jpegdegradation}, enhancing the model's robustness to diverse input conditions. Content-Aware Degradation, in particular, preserves semantically important features (e.g., edges and textures) while applying controlled distortions to less salient areas, enabling the model to generalize effectively across datasets with varying levels of detail and motion. The randomized parameters of these techniques (e.g., kernel sizes, sigma values) prevent overfitting, ensuring that the model performs reliably on noisy or low-quality inputs, as evidenced by its strong performance on the REDS dataset, which features fast-moving scenes. These degradation strategies, combined with the patch-based training system, create a robust training environment that directly contributes to the superior quantitative and qualitative results observed, positioning the framework as a versatile solution for real-world video enhancement challenges.

While this study primarily focused on $2 \times$ and $4 \times$ upscaling, the framework's modular design and scalable training pipeline make it adaptable to higher upscaling factors (e.g., $8 \times$) or other video restoration tasks, such as deblurring or denoising. The incorporation of attention mechanisms, particularly through Non-Local Blocks ~\cite{wang2018nonlocalneuralnetworks} ~\cite{vaswani2023attentionneed}, highlights their potential for capturing complex frame relationships, paving the way for further exploration of attention-based architectures in generative adversarial networks. The framework's performance approaches industry standards like NVIDIA's Deep Learning Super Sampling (DLSS) ~\cite{watson2020deeplearningtechniquessuperresolution}, offering a hardware-agnostic alternative that democratizes access to advanced video super-resolution. The development of a website to showcase the framework further enhances its practical utility, aligning with the objective of demonstrating real-world applications.

Future research can build on these achievements by addressing remaining challenges, such as optimizing computational efficiency for real-time deployment, refining metric reliability, and expanding the degradation repertoire to include codec-specific artifacts. Exploring unsupervised or self-supervised learning approaches could reduce reliance on paired HR-LR datasets, while applying the framework to niche domains like medical imaging or animated content could broaden its impact.

Overall, this work contributes valuable insights, methodologies, and a robust framework to the field of video super-resolution. By leveraging spatio-temporal feature extraction, patch-based training, and advanced data degradation techniques, it sets a new benchmark for AI-driven visual enhancement technologies, offering a foundation for future advancements that can transform how it is processed and experienced the video content.






    \noindent
\newline
{\large\textbf{List of Abbreviations}}
\begin{abbrv}
    
    \item[GAN]              Generative Adversarial Network
    \item[GT]               Ground truth
    \item[HR]               High-Resolution
    \item[LPIPS]            Learned Perceptual Image Patch Similarity
    \item[LR]               Low-Resolution
    \item[LReLU]            Leaky Rectified Linear Unit
    \item[ReLU]             Rectified Linear Unit
    \item[RRDB]             Residual-in-Residual Dense Block
    \item[SISR]             Single-Image Super Resolution
    \item[SRGAN]            Super-Resolution Generative Adversarial Network
    
\end{abbrv}

    
    
    \begin{appendices}



% ...
    \end{appendices}
    
    \bibliography{references}
    
\end{document}